\documentclass{article}

\usepackage[T1]{fontenc}
\usepackage{microtype}
\usepackage{graphicx}
\usepackage{subcaption}
\usepackage{booktabs} 

\usepackage{pdfx}
\usepackage{enumitem}

\PassOptionsToPackage{colorlinks,urlcolor=myorange,citecolor=myorange,linkcolor=myorange}{hyperref}

\usepackage[preprint]{icml2026}

\usepackage{amsmath}
\usepackage{amssymb}
\usepackage{mathtools}
\usepackage{amsthm}

\usepackage{basic_commands}

\usepackage{microtype}
\usepackage{graphicx}
\usepackage{booktabs} %
\usepackage{subcaption}
\usepackage[most]{tcolorbox}
\usepackage{float}
\usepackage{wrapfig}
\usepackage{lipsum}
\usepackage{colortbl}
\usepackage{soul}
\usepackage{enumitem}
\usepackage{multirow}
\usepackage{makecell}
\usepackage{tablefootnote}
\usepackage[para,online,flushleft]{threeparttable}
\setlist{label=\textbullet}

\theoremstyle{plain}

\theoremstyle{definition}

\theoremstyle{remark}



\usepackage{algorithm} 
\usepackage[noend]{algpseudocode}
\usepackage[endLComment=,italicComments=false]{algpseudocodex}

\usepackage{xcolor}
\definecolor{myorange}{HTML}{E67E22}

\usepackage[capitalize,noabbrev]{cleveref}

\definecolor{mypurple}{HTML}{7A4988}
\definecolor{mygreen}{HTML}{2ca02c}
\definecolor{myblue}{HTML}{1f77b4}
\hypersetup{urlcolor=myorange, citecolor=myorange, linkcolor=myorange}

\usepackage[textsize=tiny]{todonotes}

\providecommand{\ours}[1][]{{\protect\color{black}{TQL\textbf{#1}}}}
\newcommand{\primaryaffiliation}{Stanford University}
\newcommand{\primaryurl}{https://pd-perry.github.io/transformer-q-learning/}

\makeatletter
\renewcommand{\@pa}[1]{%
  \ifcsname the@affil#1\endcsname
  \else
    \ifcsname @icmlsymbol#1\endcsname
    \else
      \stepcounter{@affiliationcounter}%
      \newcounter{@affil#1}%
      \setcounter{@affil#1}{\value{@affiliationcounter}}%
    \fi
  \fi%
  \ifcsname @icmlsymbol#1\endcsname
    \textsuperscript{\csname @icmlsymbol#1\endcsname\,}%
  \fi
}

\renewcommand{\printAffiliationsAndNotice}[1]{\global\icml@noticeprintedtrue%
  {\let\thefootnote\relax\footnotetext{\hspace*{-\footnotesep}\ificmlshowauthors #1\fi%
      \ifdefined\icmlcorrespondingauthor@text
         { }Correspondence to: \icmlcorrespondingauthor@text.
      \else
        {\bf AUTHORERR: Missing \textbackslash{}icmlcorrespondingauthor.}
      \fi

      \ \\
      \Notice@String
    }
  }
}
\makeatother

\icmltitlerunning{Scaling Q-Functions with Transformers by Preventing Attention Collapse}

\begin{document}

\twocolumn[
  \icmltitle{TQL: Scaling Q-Functions with Transformers \\ by Preventing Attention Collapse}



  \icmlsetsymbol{equal}{*}
  \renewcommand{\icmlEqualContribution}{\textsuperscript{*}Equal contribution }

  \begin{icmlauthorlist}
    \icmlauthor{Perry Dong}{equal,yyy}
    \icmlauthor{Kuo-Han Hung}{equal,yyy}
    \icmlauthor{Alexander Swerdlow}{yyy}
    \icmlauthor{Dorsa Sadigh}{yyy}
    \icmlauthor{Chelsea Finn}{yyy}
  \end{icmlauthorlist}

  \vspace{0.03in}
  \centerline{\primaryaffiliation}
  \vspace{0.02in}
  \centerline{\url{\primaryurl}}

  \icmlaffiliation{yyy}{\primaryaffiliation}

  \icmlcorrespondingauthor{Perry Dong}{perryd@stanford.edu}
  \icmlcorrespondingauthor{Kuo-Han Hung}{khhung@stanford.edu}


  \vskip 0.3in
]










\printAffiliationsAndNotice{\icmlEqualContribution}

\begin{abstract}
   Despite scale driving substantial recent advancements in machine learning, reinforcement learning (RL) methods still primarily use small value functions. Naively scaling value functions -- including with a transformer architecture, which is known to be highly scalable -- often results in learning instability and \textit{worse} performance. In this work, we ask what prevents transformers from scaling effectively for value functions? Through empirical analysis, we identify the critical failure mode in this scaling: attention scores collapse as capacity increases. Our key insight is that we can effectively prevent this collapse and stabilize training by controlling the entropy of the attention scores, thereby enabling the use of larger models. To this end, we propose Transformer Q-Learning (\ours{}), a method that unlocks the scaling potential of transformers in learning value functions in RL. Our approach yields up to a 43\% improvement in performance when scaling from the smallest to the largest network sizes, while prior methods suffer from performance degradation. \looseness=-1 
\end{abstract}


\section{Introduction} \label{sec:intro}

Recent advances in deep learning have demonstrated that scaling model capacity yields substantial performance gains across domains such as natural language processing~\citep{openai2024gpt4ocard}, computer vision~\citep{siméoni2025dinov3}, and robotics~\citep{intelligence2025pi05visionlanguageactionmodelopenworld,geminiroboticsteam2025geminirobotics15pushing}. However, unlike supervised learning---where increasing parameters typically improves performance---reinforcement learning (RL) has historically faced significant scaling challenges. Larger networks often fail to better capture the data distribution and instead induce training instabilities that degrade performance. Recent works have explored using larger policy networks in RL. In this work, we focus on scaling up value functions. As policies are ultimately derived from value functions, increased parameterization of the value function should substantially benefit RL performance; yet, using larger value function networks often results in \emph{worse} performance, and how to effectively scale them remains an open question. In traditional supervised learning, transformers have been demonstrated to scale well across numerous domains. Therefore, we ask: what prevents transformers from effectively scaling in RL value learning?


\begin{figure*}[t!]
    \centering
    \includegraphics[width=0.99\textwidth]{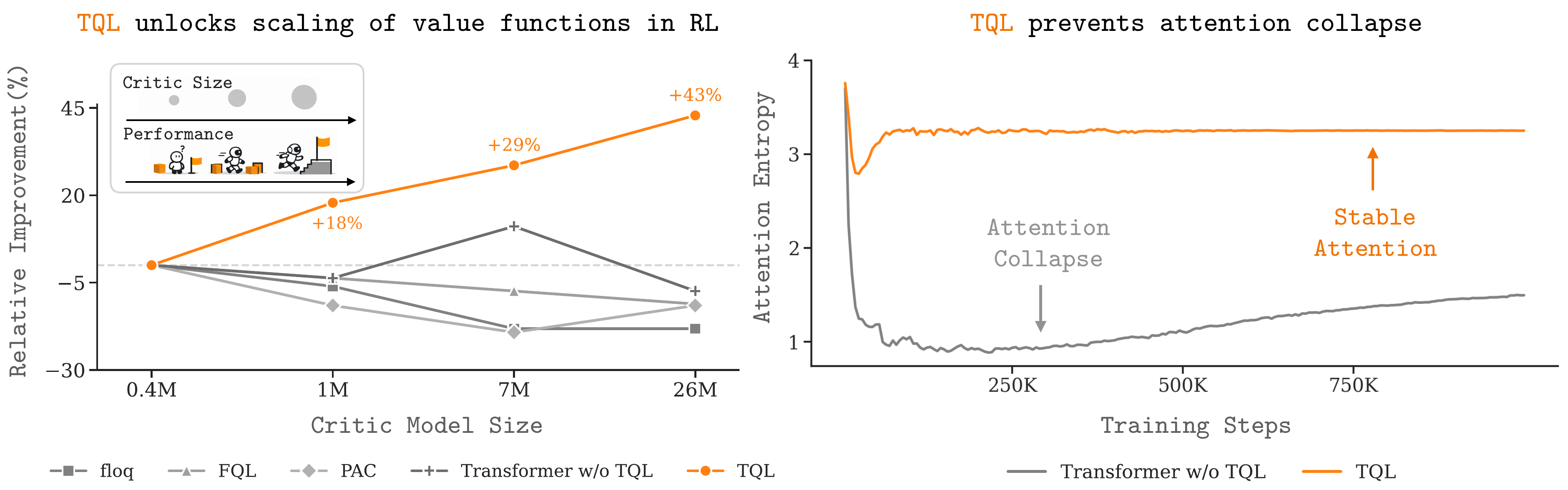}
    \caption{
        \small 
        \textbf{\ours{} unlocks scaling of value functions in RL by preventing attention collapse.} Left: Scaling results of \ours{} compared with prior approaches with different generative model backbones. The results are reported as relative improvement over the smallest model size, averaged across seeds and all 25 tasks in our evaluation suite. \ours{} increases in performance as network size increases, while prior approaches are not able to effectively use the extra capacity and drop in performance. Right: Attention entropy with and without \ours{}, averaged across 25 tasks. \ours{} scales effectively by preventing attention collapse that occurs with scaling up the transformer architecture for value function training. 
    }
    \vspace{-0.5cm}
    \label{fig:scale}
\end{figure*}



Popular prior approaches to scaling up networks in RL employ periodic network resets to maintain plasticity under higher update-to-data (UTD) ratios during online training~\citep{nauman2024biggerregularizedoptimisticscaling,schwarzer2023bigger}. While these methods preserve learning capacity, they are compute-heavy as the network needs to be retrained from scratch and they risk catastrophic forgetting through parameter reinitialization that can make online learning unsafe. Alternative approaches have explored normalization techniques~\citep{lee2025simbasimplicitybiasscaling} or alternative architectures~\citep{obandoceron2024mixturesexpertsunlockparameter}, which often require specific architectural changes, making it challenging to apply to all settings such as in the case of pretrained networks. In this work, we focus on understanding why transformers do not scale for value function learning in RL, and propose a general, minimal framework to address these issues and enable scaling.


To this end, we propose Transformer Q-Learning (\ours{}), a simple framework for scalable value function training in RL. \ours{} leverages the scalability of the transformer architecture for value function training, while keeping architectural changes as minimal as possible. Directly applying transformers to value learning yields severe pathologies (\cref{sec:scale}). Through empirical analysis, we identify the critical failure mode in scaling up transformer value function training: transformers collapse their attention weights as capacity increases, attending predominantly to a handful of tokens while ignoring the rest, and this results in larger models struggling to learn appropriate attention patterns during value bootstrapping. As a simple remedy, we propose per-layer control of the entropy of the attention scores. By adjusting the entropy of attention scores toward a target value, \ours{} ensures that the model effectively distributes attention across all input tokens to enable stable training by preventing collapse. This enables the larger model to use its capacity to fit useful signals and unlock scaling of transformers for value function training (\cref{fig:scale}).

Our main contribution is \ours{}, a framework for guiding the attention entropy as a way to provide stable, effective value function training as the model scales up in capacity. We analyze \ours{} in an offline RL setting and evaluate on challenging continuous control tasks from the OGBench benchmark~\citep{park2025ogbench}, demonstrating the effectiveness of \ours{} in scaling up value functions compared to prior approaches.


\section{Related Work} \label{sec:related_work}

\begin{figure*}[t!]
    \centering
    \includegraphics[width=0.99\linewidth]{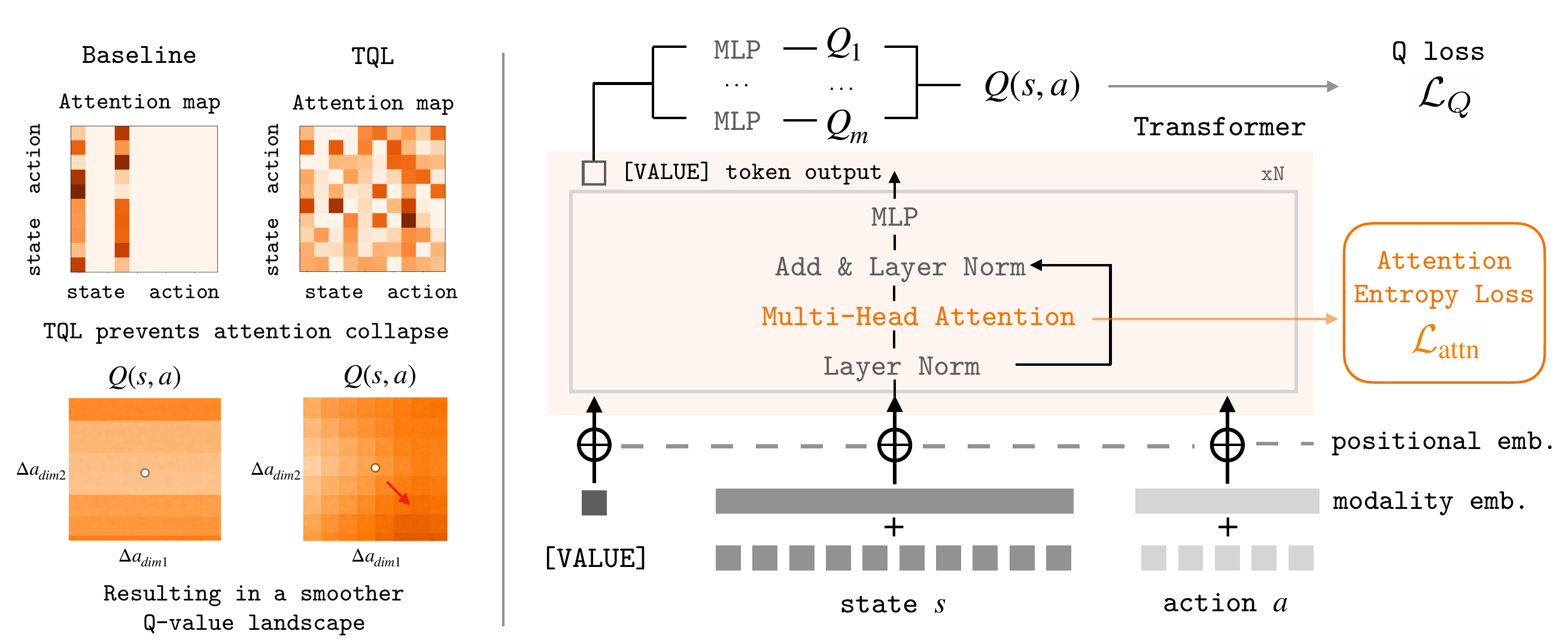}
    \caption{
        \small 
        \textbf{Scaling value networks with \ours{}.} \ours{} prevents attention collapse in larger networks by controlling the target entropy of attention scores (right). Compared to an unregularized model, \ours{} exhibits a more uniform attention distribution and a smoother value landscape (left). 
    }
    \vspace{-0.2cm}
    \label{fig:tql}
\end{figure*}

\textbf{Offline RL. } The goal of offline RL~\citep{levine2020offlinereinforcementlearningtutorial} is to train a policy on an offline dataset without environment interactions. The central concern of offline RL is handling actions that are out of the dataset distribution, where estimates of value are unreliable. There are two primary approaches to address this issue. One approach is to use conservatism to learn a pessimistic value function to penalize out-of-distribution actions~\citep{tarasov2023revisitingminimalistapproachoffline, kumar2020conservative, wu2019behaviorregularizedofflinereinforcement,wu2021uncertainty, yu2020mopo}. Another is to apply policy constraints to keep the policy actions close to the behavior distribution~\citep{dong2025valueflows,park2025flowqlearning,hansenestruch2023idqlimplicitqlearningactorcritic,kostrikov2021offline,fujimoto2021minimalist,nair2021awac,kidambi2020morel,fujimoto2019off}. Recent works have also explored modern generative architectures for value function learning~\citep{dong2025valueflows,agrawalla2025floqtrainingcriticsflowmatching}. Our work also proposes a novel value function design, but with an explicit focus on scalability and enabling effective performance improvements as model capacity increases. The techniques we introduce are largely orthogonal to these prior works as \ours{}, in principle, can be combined with these different algorithms and distribution choices.

\textbf{Transformers in RL. } Transformers have demonstrated significant success in RL in trajectory modeling~\citep{wu2023elasticdecisiontransformer,zheng2022onlinedecisiontransformer,furuta2022generalizeddecisiontransformeroffline,chen2021decision,janner2021offlinereinforcementlearningbig}, world modeling~\citep{fan2024scaling}, and joint policy-value architectures~\citep{springenberg2024offlineactorcriticreinforcementlearning}. In contrast, we focus on the question of how to effectively scale transformers specifically for value functions. While there have been some works that explore designing transformer-based value functions~\citep{obandoceron2024mixturesexpertsunlockparameter, chebotar2023qtransformer}, the reason transformers do not scale directly for value functions remains unclear, and we lack a reliable recipe for large-scale transformer-based value function training. ~\citet{intelligence2025pi06vlalearnsexperience} explores training value functions with Monte Carlo returns, whereas we explore how to train such value functions with bootstrapping, which has been shown to work better but often has more training instabilities. Various other works have examined training instabilities that arise from scaling transformers in other domains such as NLP~\citep{zhai2023stabilizing, wortsman2023smallscale, lai2024hybridnorm}. In contrast to these approaches, we focus exclusively on value function modeling with transformers, investigating architectural choices that enable stable and effective scaling in a value function training setting. 

\textbf{Scaling Parameters in RL. } Scaling parameter sizes in RL remains challenging compared to supervised learning domains. Existing approaches primarily address this through regularization. One line of work employs periodic network resets to maintain plasticity during training~\citep{nauman2024biggerregularizedoptimisticscaling,schwarzer2023bigger}, though this is compute-heavy as the
network needs to be periodically retrained from scratch and it risks catastrophic forgetting of previously learned actions, which can make online training in the real world unsafe. Other approaches have explored architectural and objective modifications through normalization techniques to stabilize training with larger networks~\citep{lee2025simbasimplicitybiasscaling}, using specific architectures such as mixture-of-experts or multi-skip residual connections~\citep{obandoceron2024mixturesexpertsunlockparameter, castanyer2025stablegradientsstablelearning}, or using categorical losses instead of regression for learning the value function~\citep{farebrother2024stop}. Recent work has examined scaling network depth for both actor and critic networks~\citep{wang20251000layernetworksselfsupervised}, demonstrating improvements in self-supervised contrastive RL settings. While these methods address specific scaling challenges, they are often limiting and require specific settings or changes to the architecture, making them unable to be applied in all cases such as in the case of pretrained models. In this work, we focus on designing a simple and general method for scaling up transformers for value function training in RL.




\section{Preliminaries} \label{sec:prelim}

We consider a Markov decision process (MDP) defined by a tuple $\{\mathcal{S}, \mathcal{A}, \rho, r, \gamma, T\}$, where $\mathcal{S}$ is the state space, $\mathcal{A}$ is the action space, $\rho(s)$ is the initial state distribution, $r:\mathcal{S}\times\mathcal{A}\rightarrow\mathbb{R}$ is a function defining the rewards, $\gamma\in[0,1]$ is a discount factor, and $T(s'|s,a)$ specifies the transition probability. The goal of RL is to learn a policy that maximizes the expected sum of discounted returns $\mathbb{E}_{\pi}[\sum_{t=0}^{T}\gamma^t r(s_t, a_t)]$. In this paper, we consider the problem of offline RL, which additionally assumes a fixed dataset $\mathcal{D}=\{(s,a,r,s')\}$ containing pre-collected transitions. The aim of offline RL is to maximize returns by learning from the fixed dataset without environment interactions.

\section{Transformer Q-Learning} \label{sec:method}

In this section, we present the framework of \ours{} to enable scalable value function training with transformer architectures. We identify the key challenge: naively scaling up transformers results in a collapse of the entropy of attention weights as the network struggles to learn which tokens to attend to with its extra capacity (full analysis in \cref{sec:scale}). Our key insight is that preventing this collapse of entropy enables transformers to scale effectively for value function training, and we can do so by directly controlling the entropy of the attention scores towards a target value. We first describe our setup of training value functions with transformers (\cref{sec:setup}), then present the approach for controlling attention score entropy (\cref{sec:entropy_tuning}), and finally discuss implementation details, specifically the incorporation of learnable modality embeddings to help larger models better attend to task-relevant patterns and policy extraction (\S\ref{sec:details}). The complete \ours{} framework is illustrated in \cref{fig:tql} and the full algorithm is described in~\cref{alg:ours}. 


\subsection{Transformer-based Value Functions}
\label{sec:setup}

We use a transformer architecture for learning the Q-function $Q(s, a)$ that maps state-action pairs to expected returns. We use a standard transformer decoder with full self-attention to predict value, where we treat each dimension of the state $s \in \mathbb{R}^{n_s}$ and action $a \in \mathbb{R}^{n_a}$ as a token and project each scalar to the hidden dimension. The tokens are combined with positional encodings before being processed by a stack of transformer layers. We prepend a learnable \texttt{[VALUE]} token to the sequence, whose final representation is passed through an MLP head to produce the value prediction. We use the standard Q-learning objective 
\begin{equation}
    \mathcal{L}_{Q}(\phi) = \mathbb{E}[(Q_\phi(s, a) - r - \gamma Q_{\phi'}(s', a'))^2] \label{eq:critic}
\end{equation}
where $Q_{\phi'}$ is a delayed copy of $Q_{\phi}$. This design enables the model to learn complex dependencies between state and action tokens through self-attention mechanisms while maintaining simplicity. As we demonstrate in ~\cref{sec:scale}, simply scaling the parameter capacity of this standard design leads to training instabilities and actually leads to \textit{worse} performance, rather than improved performance. In fact, we find that scaling results in a predictable decrease in performance -- in one case (\cref{fig:ana}), the average success rate plummets from $46\%$ to just $6\%$.

\subsection{Directly Guiding Entropy of Attention Scores}
\label{sec:entropy_tuning}


We hypothesize that the poor performance of large transformer value networks can be attributed to \emph{attention collapse}. Specifically, our experiments in \cref{sec:scale} show that the attention distributions of large Q-networks become increasingly concentrated on a handful of tokens. Once collapsed, the model is unable to estimate accurate values of expected discounted returns and obtain high performance. 


To address the challenges of unstable attention, we can directly guide the entropy of attention distributions during training toward a target value. For a given attention layer $\ell$, let $A^{\ell} \in \mathbb{R}^{n \times n}$ denote the attention score matrix (after softmax) of the layer, where $n = 1 + n_s + n_a$ is the total sequence length, $n_s$ is the dimension of the state, $n_a$ is the dimension of the actions, and 1 corresponds to the \texttt{[VALUE]} token. The entropy of the attention distribution for the $i$-th query token is:
\begin{equation}
H_i^{\ell} = -\sum_{j=1}^{n} A_{ij}^{\ell} \log A_{ij}^{\ell}
\end{equation}

Inspired by maximum entropy RL~\citep{haarnoja2018softactorcriticoffpolicymaximum}, we introduce a learnable temperature parameter $\alpha$ and optimize it to maintain a target entropy $\bar{H}$:
\begin{align}
\mathcal{L}_{\text{temp}} (\alpha) &= \frac{1}{L}\sum_{\ell=1}^{L}\alpha^{\ell} \left( H^{\ell} - \bar{H} \right) \label{eq:temp} \\
\mathcal{L}_{\text{attn}}(\phi) &= -\frac{1}{L}\sum_{\ell=1}^{L} \alpha^{\ell} H^{\ell} \label{eq:attn}
\end{align}

where $H^{\ell}=\frac{1}{n}\sum_{i=1}^{n} H_{i}^{\ell}$. In practice, we parameterize $\alpha$ as an exponential $\exp(\hat{\alpha})$ and optimize $\hat{\alpha}$. By maximizing attention entropy in the attention loss, the network is encouraged to diversify its attention to focus on all of the input tokens, while ensuring stability as the temperature term adjusts entropy toward a desired level. The temperature parameter is optimized jointly with the network parameters via gradient descent to stabilize the attention entropy:
\begin{equation} \label{formula:critic_loss}
\mathcal{L}_{\text{critic}}(\phi,\alpha) = \mathcal{L}_Q(\phi) + \mathcal{L}_{\text{attn}}(\phi) + \mathcal{L}_{\text{temp}}(\alpha)
\end{equation}

%
\label{sec:layerwise_tuning} 
We make two important design choices in how the attention entropy control is applied. First, we use \emph{layer-wise} temperature parameters $\alpha^{\ell}$ rather than a single global parameter. Different layers in a transformer can learn to attend to different patterns. By enabling each layer to maintain its own target entropy, we allow the model to learn the separate attention patterns for each layer that are appropriate for value function learning. Second, we apply a separate temperature $\alpha_{\texttt{[VALUE]}}^{\ell}$ for controlling the attention entropy of the \texttt{[VALUE]} token, which aggregates information from all other tokens to produce the final Q-value.  
The \texttt{[VALUE]} token can contain different attention entropy patterns, as it is responsible for aggregating information from all state and action tokens to produce an accurate value estimate. These design choices allow each layer and the \texttt{[VALUE]} token to independently maintain appropriate entropy levels, leading to stable training. 

The automatic entropy control mechanism allows \ours{} to prevent entropy collapse by directly adjusting the entropy to a desired value, as the temperature coefficient adaptively regulates the entropy in attention distributions to extract information from all tokens, while still allowing the model to learn which tokens are most relevant for value prediction. This enables larger models to utilize their extra capacity to learn the most useful signals, instead of collapse. This approach introduces an additional hyperparameter $\bar{H}$ for the target entropy values. In practice, we use a similar set of values for $\bar{H}$ across tasks and environments, and provide recommendations for how to select the value of this hyperparameter in~\cref{sec:appendix_detail}.



\subsection{Implementation Details}
\label{sec:details}

\paragraph{Learnable Modality Embedding. }  \label{sec:modality_embedding} Beyond stabilizing the entropy of attention scores, we seek to more effectively utilize the capacity of larger models. To this end, we introduce learnable modality embeddings $e_s$ and $e_a$ that are added to state and action tokens respectively. These modality embeddings provide a simple mechanism for the model to differentiate between state and action information. Combined with positional embeddings, this allows larger models to focus the attention on states or actions depending on what information is most relevant for Q-value prediction.

\paragraph{Policy Extraction. } \ours{} can, in principle, be applied to any value-based policy extraction scheme and has no method-specific design decisions. We choose a recent, high-performing offline RL policy extraction scheme~\citep{park2025flowqlearning}: we learn a one-step flow policy $\pi_{\omega}$ that is behavior constrained through distillation toward a Behavior Cloning (BC) flow policy $\pi^{\beta}$ while steering it to maximize Q-values:

\begin{align}
    \gL&_{\tiny\mathrm{BC}}(\theta) \!=\! \E_{\tiny \substack{s,a\!=\!x^1\!\sim\!\gD, \\ x^0\!\sim\!\gN(0,I_d), \\ t\!\sim\!\mathrm{Unif}([0,1])}}
    \!\left[
        \|\pi^{\beta}_\theta(t,s,x^t) \!-\! (x^1 \!-\! x^0)\|_2^2
    \right]\!, \label{eq:bc-flow-policy-obj} \raisetag{1.3\normalbaselineskip}
    \\
    \gL&_{{\text{\tiny OS}}}(\omega) \!=\!
    \underbrace{\E_{\tiny s\sim\gD, a^\pi\sim\pi_\omega}\![-Q_\phi(s,a^\pi)]}_{\text{\tiny Q}}
    \!+\! \underbrace{\alpha \gL_{\tiny\mathrm{Distill}}(\omega)}_{\text{\tiny BC}}, \label{eq:one-step-flow-policy-obj}
    \\
    \gL&_{\tiny\mathrm{Distill}}(\omega) \!=\!
    \E_{\tiny\substack{s\sim\gD, \\ z\sim\gN(0,I_d)}}
    \!\left[\|\mu_\omega(s,z) \!-\! \mu_\theta(s,z)\|_2^2\right]\!. \label{eq:distill}
\end{align}


where $\alpha$ controls the strength of the behavior constraint on the policy. This allows the policy to be behavior-constrained while still maximizing value.

\begin{algorithm}[t]
\caption{\textcolor{myorange}{Transformer Q-Learning (TQL)}}
\label{alg:ours}
\begin{algorithmic}

\State \textbf{Input:} Offline dataset $\gD$, \textcolor{myorange}{target entropy coefficient $\bar H$}, number of layers $L$
\State Initialize Q-network $Q_\phi$ with transformer architecture
\State Initialize target network $Q_{\phi'}$ with $\phi' \leftarrow \phi$
\State Initialize policy network $\pi_{\omega}$, $\pi^{\beta}_{\theta}$, temperature parameters $\alpha=\{\alpha^{\ell},\alpha_{\texttt{[VALUE]}}^{\ell}\}_{{\ell}=1}^{L}$

\vspace{5pt}

\For{each training iteration}
    \State Sample batch $(s, a, r, s') \sim \gD$

    \BeginBox[fill=myorange!8]
        \LComment{\textcolor{myorange}{Train Critic}}
        \State Sample next actions $a' \sim \pi_{\theta}(a'|s')$

        \State Compute TD loss $\mathcal{L}_Q$ with Equation (1)

        
        \State Compute \textcolor{myorange}{attention entropies $\{H^{\ell}, H_{ \texttt{[VALUE]}}^{\ell}\}_{{\ell}=1}^{L}$} 
        
        \State Compute \textcolor{myorange}{entropy losses $\mathcal{L}_{\text{attn}}$} with Equation (4)
        
        \State Train critic $Q_\phi$ by minimizing $\mathcal{L}_Q + \textcolor{myorange}{\mathcal{L}_{\text{attn}}}$
        
        \State Train temperatures $\alpha$ by minimizing \textcolor{myorange}{temperature loss $\mathcal{L}_{\text{temp}}$} with Equation (3)


        
        \State Periodically update target: $\phi' \leftarrow \tau \phi + (1-\tau)\phi'$
    \EndBox
    
    \vspace{-10pt}
    
    \BeginBox[fill=white]
        \LComment{Policy Extraction}
        \State Update one-step policy $\pi_\omega$ with $\mathcal{L}_{\text{OS}}$ in Equation (7)
        \State Update flow policy $\pi_{\theta}^{\beta}$ with $\mathcal{L}_{\text{BC}}$ in Equation (6)
    \EndBox
    \vspace{-10pt}
    
\EndFor

\State \textbf{return} $Q_\phi, \pi_\omega$

\end{algorithmic}
\end{algorithm}

\section{Experiments} \label{sec:experiment}
\begin{figure*}[t!]
    \centering
    \includegraphics[width=0.99\linewidth]{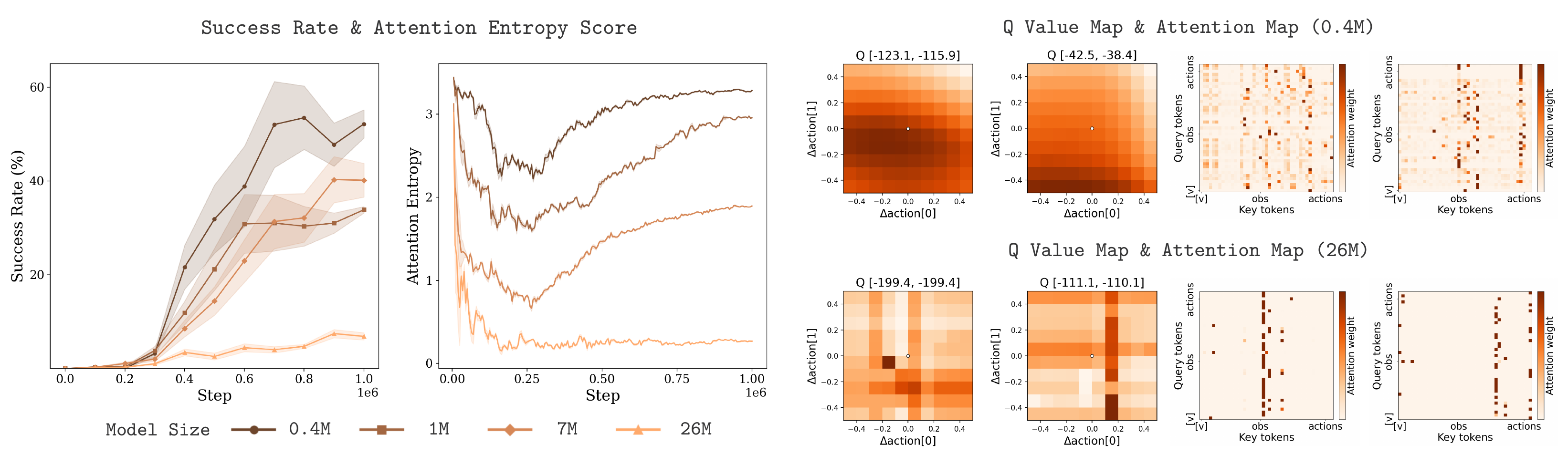}
    \caption{
        \small 
        \textbf{Scaling transformers for value functions results in entropy collapse and worse performance. } Left: Visualizations of the success rate and attention entropy of the transformer model without \ours{} across five \texttt{cube-double} tasks under different model sizes. Right: Q-value landscapes and attention maps for the smallest (0.4M) and largest (26M) models. The larger transformer learns highly non-smooth value surfaces and exhibits high-frequency oscillations and discontinuities that are absent in its smaller counterpart.
    }
    \label{fig:ana}
    \vspace{-0.2cm}
\end{figure*}

In this section, we empirically evaluate \ours{} compared to prior RL methods on a variety of challenging tasks, focusing on scaling and performance. The goal of our experiments is to answer the following key questions:

\begin{enumerate}[start=1,label={(\bfseries Q\arabic*)}]
\vspace{1cm}
    \item What prevents transformers from scaling well for value functions?
    \item Does \ours{} improve transformer scalability in a value function setting compared to prior approaches?
    \item Can \ours{} effectively learn a policy from a static offline dataset?
    \item What are the most important components of \ours{}? 
\end{enumerate}


\subsection{Experiment Setup}
\begin{table*}[h!]
\centering
\setlength{\tabcolsep}{6pt}
\scalebox{0.72}
{
\begin{tabular}{lcccccccccc}
\toprule
& \multicolumn{6}{c}{MLP Q} & \multicolumn{1}{c}{Flow Q} & \multicolumn{3}{c}{Transformer Q} \\
\cmidrule(lr){2-7} \cmidrule(lr){8-8} \cmidrule(lr){9-11}
& BC & IQL & ReBRAC & FBRAC & IFQL & FQL & floq & Q-T & PAC & \textcolor{myorange}{TQL} \\
\midrule
  \texttt{cube-double-play-singletask-task1-v0}  & $8 \pm 3$ & $27 \pm 5$ & $45 \pm 6$ & $47 \pm 11$ & $35 \pm 9$ & $61 \pm 9$ & $45 \pm 23$ & $2 \pm 1$ & $28 \pm 16$ & $\mathbf{95 \pm 2}$  \\
  \texttt{cube-double-play-singletask-task2-v0}  & $0 \pm 0$ & $1 \pm 1$ & $7 \pm 3$ & $22 \pm 12$ & $9 \pm 5$ & $36 \pm 6$ & $42 \pm 23$ & $0 \pm 0$ & $36 \pm 16$ & $\mathbf{84 \pm 8}$  \\
  \texttt{cube-double-play-singletask-task3-v0}  & $0 \pm 0$ & $0 \pm 0$ & $4 \pm 1$ & $4 \pm 2$ & $8 \pm 5$ & $22 \pm 5$ & $\mathbf{59 \pm 15}$ & $0 \pm 0$ & $50 \pm 11$ & $\mathbf{60 \pm 27}$  \\
  \texttt{cube-double-play-singletask-task4-v0}  & $0 \pm 0$ & $0 \pm 0$ & $1 \pm 1$ & $0 \pm 1$ & $1 \pm 1$ & $5 \pm 2$ & $13 \pm 8$ & $0 \pm 0$ & $10 \pm 6$ & $\mathbf{14 \pm 6}$  \\
  \texttt{cube-double-play-singletask-task5-v0}  & $0 \pm 0$ & $4 \pm 3$ & $4 \pm 2$ & $2 \pm 2$ & $17 \pm 6$ & $19 \pm 10$ & $46 \pm 11$ & $0 \pm 0$ & $25 \pm 21$ & $\mathbf{84 \pm 9}$  \\
  \midrule
  \texttt{cube-double-play (average)}  & $2 \pm 1$ & $6 \pm 2$ & $12 \pm 3$ & $15 \pm 6$ & $14 \pm 5$ & $29 \pm 6$ & $41 \pm 16$ & $0 \pm 0$ & $30 \pm 14$ & $\mathbf{67 \pm 10}$  \\
\midrule
  \texttt{scene-play-singletask-task1-v0}  & $19 \pm 6$ & $94 \pm 3$ & $\mathbf{95 \pm 2}$ & $\mathbf{96 \pm 8}$ & $\mathbf{98 \pm 3}$ & $\mathbf{100 \pm 0}$ & $\mathbf{100 \pm 0}$ & $27 \pm 6$ & $\mathbf{100 \pm 0}$ & $\mathbf{100 \pm 0}$  \\
  \texttt{scene-play-singletask-task2-v0}  & $1 \pm 1$ & $12 \pm 3$ & $50 \pm 13$ & $46 \pm 10$ & $0 \pm 0$ & $76 \pm 9$ & $\mathbf{92 \pm 9}$ & $6 \pm 3$ & $54 \pm 16$ & $81 \pm 14$  \\
  \texttt{scene-play-singletask-task3-v0}  & $1 \pm 1$ & $32 \pm 7$ & $55 \pm 16$ & $78 \pm 4$ & $54 \pm 19$ & $\mathbf{98 \pm 1}$ & $\mathbf{97 \pm 4}$ & $1 \pm 1$ & $\mathbf{98 \pm 2}$ & $\mathbf{98 \pm 2}$  \\
  \texttt{scene-play-singletask-task4-v0}  & $2 \pm 2$ & $0 \pm 1$ & $3 \pm 3$ & $4 \pm 4$ & $0 \pm 0$ & $5 \pm 1$ & $4 \pm 4$ & $1 \pm 1$ & $4 \pm 3$ & $\mathbf{25 \pm 17}$  \\
  \texttt{scene-play-singletask-task5-v0}  & $\mathbf{0 \pm 0}$ & $\mathbf{0 \pm 0}$ & $\mathbf{0 \pm 0}$ & $\mathbf{0 \pm 0}$ & $\mathbf{0 \pm 0}$ & $\mathbf{0 \pm 0}$ & $\mathbf{0 \pm 0}$ & $\mathbf{0 \pm 1}$ & $\mathbf{0 \pm 0}$ & $\mathbf{0 \pm 0}$  \\
  \midrule
  \texttt{scene-play (average)}  & $5 \pm 2$ & $28 \pm 3$ & $41 \pm 7$ & $45 \pm 5$ & $30 \pm 4$ & $56 \pm 2$ & $\mathbf{59 \pm 3}$ & $7 \pm 2$ & $51 \pm 4$ & $\mathbf{61 \pm 7}$  \\
\midrule
  \texttt{puzzle-3x3-play-singletask-task1-v0}  & $5 \pm 2$ & $33 \pm 6$ & $\mathbf{97 \pm 4}$ & $63 \pm 19$ & $\mathbf{94 \pm 3}$ & $90 \pm 4$ & $\mathbf{94 \pm 5}$ & $3 \pm 2$ & $92 \pm 4$ & $\mathbf{96 \pm 4}$  \\
  \texttt{puzzle-3x3-play-singletask-task2-v0}  & $1 \pm 1$ & $4 \pm 3$ & $1 \pm 1$ & $2 \pm 2$ & $1 \pm 2$ & $16 \pm 5$ & $\mathbf{22 \pm 9}$ & $1 \pm 1$ & $17 \pm 12$ & $19 \pm 12$  \\
  \texttt{puzzle-3x3-play-singletask-task3-v0}  & $1 \pm 1$ & $3 \pm 2$ & $3 \pm 1$ & $1 \pm 1$ & $0 \pm 0$ & $10 \pm 3$ & $\mathbf{16 \pm 5}$ & $0 \pm 1$ & $8 \pm 4$ & $7 \pm 5$  \\
  \texttt{puzzle-3x3-play-singletask-task4-v0}  & $1 \pm 1$ & $2 \pm 1$ & $2 \pm 1$ & $2 \pm 2$ & $0 \pm 0$ & $\mathbf{16 \pm 5}$ & $\mathbf{16 \pm 10}$ & $0 \pm 0$ & $14 \pm 3$ & $10 \pm 4$  \\
  \texttt{puzzle-3x3-play-singletask-task5-v0}  & $1 \pm 0$ & $3 \pm 2$ & $5 \pm 3$ & $2 \pm 2$ & $0 \pm 0$ & $16 \pm 3$ & $25 \pm 7$ & $0 \pm 1$ & $9 \pm 3$ & $\mathbf{30 \pm 10}$  \\
  \midrule
  \texttt{puzzle-3x3-play (average)}  & $2 \pm 1$ & $9 \pm 3$ & $22 \pm 2$ & $14 \pm 5$ & $19 \pm 1$ & $30 \pm 4$ & $\mathbf{35 \pm 7}$ & $1 \pm 1$ & $28 \pm 5$ & $32 \pm 7$  \\
\midrule
  \texttt{puzzle-4x4-play-singletask-task1-v0}  & $1 \pm 1$ & $12 \pm 2$ & $26 \pm 4$ & $32 \pm 9$ & $49 \pm 9$ & $34 \pm 8$ & $\mathbf{60 \pm 8}$ & $0 \pm 0$ & $14 \pm 7$ & $\mathbf{59 \pm 11}$  \\
  \texttt{puzzle-4x4-play-singletask-task2-v0}  & $0 \pm 0$ & $7 \pm 4$ & $12 \pm 4$ & $5 \pm 3$ & $4 \pm 4$ & $16 \pm 5$ & $\mathbf{22 \pm 5}$ & $0 \pm 1$ & $13 \pm 5$ & $19 \pm 6$  \\
  \texttt{puzzle-4x4-play-singletask-task3-v0}  & $0 \pm 0$ & $9 \pm 3$ & $15 \pm 3$ & $20 \pm 10$ & $\mathbf{50 \pm 14}$ & $18 \pm 5$ & $36 \pm 7$ & $0 \pm 0$ & $8 \pm 7$ & $43 \pm 5$  \\
  \texttt{puzzle-4x4-play-singletask-task4-v0}  & $0 \pm 0$ & $5 \pm 2$ & $10 \pm 3$ & $5 \pm 1$ & $\mathbf{21 \pm 11}$ & $11 \pm 3$ & $19 \pm 4$ & $0 \pm 0$ & $3 \pm 2$ & $\mathbf{20 \pm 3}$  \\
  \texttt{puzzle-4x4-play-singletask-task5-v0}  & $0 \pm 0$ & $4 \pm 1$ & $7 \pm 3$ & $2 \pm 2$ & $2 \pm 2$ & $7 \pm 3$ & $14 \pm 3$ & $0 \pm 0$ & $5 \pm 3$ & $\mathbf{16 \pm 5}$  \\
  \midrule
  \texttt{puzzle-4x4-play (average)}  & $0 \pm 0$ & $7 \pm 2$ & $14 \pm 3$ & $13 \pm 5$ & $25 \pm 8$ & $17 \pm 5$ & $\mathbf{30 \pm 5}$ & $0 \pm 0$ & $9 \pm 5$ & $\mathbf{31 \pm 6}$  \\
\midrule
  \texttt{cube-triple-play-singletask-task1-v0}  & $1 \pm 1$ & $4 \pm 4$ & $1 \pm 2$ & $0 \pm 0$ & $2 \pm 2$ & $20 \pm 6$ & $17 \pm 11$ & $0 \pm 0$ & $1 \pm 1$ & $\mathbf{29 \pm 16}$  \\
  \texttt{cube-triple-play-singletask-task2-v0}  & $0 \pm 0$ & $0 \pm 0$ & $0 \pm 0$ & $0 \pm 0$ & $0 \pm 0$ & $\mathbf{1 \pm 2}$ & $0 \pm 0$ & $0 \pm 0$ & $0 \pm 0$ & $0 \pm 1$  \\
  \texttt{cube-triple-play-singletask-task3-v0}  & $0 \pm 0$ & $0 \pm 0$ & $0 \pm 0$ & $0 \pm 0$ & $0 \pm 0$ & $0 \pm 0$ & $1 \pm 1$ & $0 \pm 0$ & $0 \pm 0$ & $\mathbf{5 \pm 4}$  \\
  \texttt{cube-triple-play-singletask-task4-v0}  & $\mathbf{0 \pm 0}$ & $\mathbf{0 \pm 0}$ & $\mathbf{0 \pm 0}$ & $\mathbf{0 \pm 0}$ & $\mathbf{0 \pm 0}$ & $\mathbf{0 \pm 0}$ & $\mathbf{0 \pm 0}$ & $\mathbf{0 \pm 0}$ & $\mathbf{0 \pm 0}$ & $\mathbf{0 \pm 0}$  \\
  \texttt{cube-triple-play-singletask-task5-v0}  & $0 \pm 0$ & $\mathbf{1 \pm 1}$ & $0 \pm 0$ & $0 \pm 0$ & $0 \pm 0$ & $0 \pm 0$ & $0 \pm 0$ & $0 \pm 0$ & $0 \pm 0$ & $0 \pm 0$  \\
  \midrule
  \texttt{cube-triple-play (average)}  & $0 \pm 0$ & $1 \pm 1$ & $0 \pm 0$ & $0 \pm 0$ & $0 \pm 0$ & $4 \pm 2$ & $4 \pm 2$ & $0 \pm 0$ & $0 \pm 0$ & $\mathbf{7 \pm 4}$  \\
\midrule
  \textbf{\texttt{Average across all tasks}}  & $2 \pm 1$ & $10 \pm 2$ & $18 \pm 3$ & $17 \pm 4$ & $18 \pm 4$ & $27 \pm 4$ & $34 \pm 7$ & $2 \pm 1$ & $24 \pm 6$ & $\mathbf{40 \pm 7}$  \\
\bottomrule
\end{tabular}
}
\vspace{0.3cm}
\caption{\textbf{OGBench evaluation results.} \ours{} achieves the highest average performance on 4 out of 5 domains, as well as the best average performance across all 25 tasks. The reported scores are averaged over 3 seeds, where bold values indicate performance within 95\% of the best result per task. \looseness=-1}
\label{tab:offline-eval}
\vspace{-0.7cm}
\end{table*}

We evaluate \ours{} on the recently proposed OGBench benchmark~\citep{park2025ogbench}. Success on OGBench requires offline RL algorithms to reason over extended temporal horizons, learn effectively from unstructured data, and handle delayed credit assignment across complex, multi-stage interactions, making it a rigorous testbed for evaluating offline RL algorithms. We adopt the reward-based single-task variants (denoted as “singletask”) as we are in the setting of standard reward-maximizing RL. Our evaluation spans five domains of five distinct tasks each: \texttt{cube-double}, \texttt{cube-triple}, \texttt{scene}, \texttt{puzzle-3×3}, and \texttt{puzzle-4×4}, resulting in a total of 25 tasks. We illustrate the domains in Figure~\ref{fig:benchmark}.

\textbf{Comparisons. } In our experiments, we compare against 9 offline RL methods selected to represent a broad range of algorithms and modeling strategies, including methods that use flow-matching-based and transformer-based value functions. For standard offline RL comparisons using Gaussian policies, we include Behavior Cloning (BC), Implicit Q Learning (IQL)~\citep{kostrikov2021offline}, and ReBRAC~\citep{tarasov2023revisitingminimalistapproachoffline} as widely adopted and competitive representatives. To capture more recent state-of-the-art advances, we further evaluate methods with flow policies in offline RL, including FBRAC which is a variant of Behavior Regularized Actor Critic~\citep{wu2019behaviorregularizedofflinereinforcement} with flow policies and IFQL, which is a variant of Implicit Diffusion Q-Learning with flow policies~\citep{hansenestruch2023idqlimplicitqlearningactorcritic}, and Flow Q-Learning (FQL)~\citep{park2025flowqlearning} which uses a one-step flow policy to maximize the Q-value. Because our method focuses on value learning, we include baselines with flow-based value learning (floq~\citep{agrawalla2025floqtrainingcriticsflowmatching}) and transformer-based value learning (Q-Transformer (Q-T)~\citep{chebotar2023qtransformer} and Perceiver Actor Critic (PAC)~\citep{springenberg2024offlineactorcriticreinforcementlearning}). For both PAC and Q-T, we follow their original value function architectures and training setups, while adopting the same FQL based policy extraction procedure as our method to ensure a fair and controlled comparison. For detailed baseline implementations and hyperparameters, we provide them in Appendix~\ref{appendix:exp_details}.
\begin{figure*}[t!]
    \centering
    \includegraphics[width=0.99\linewidth]{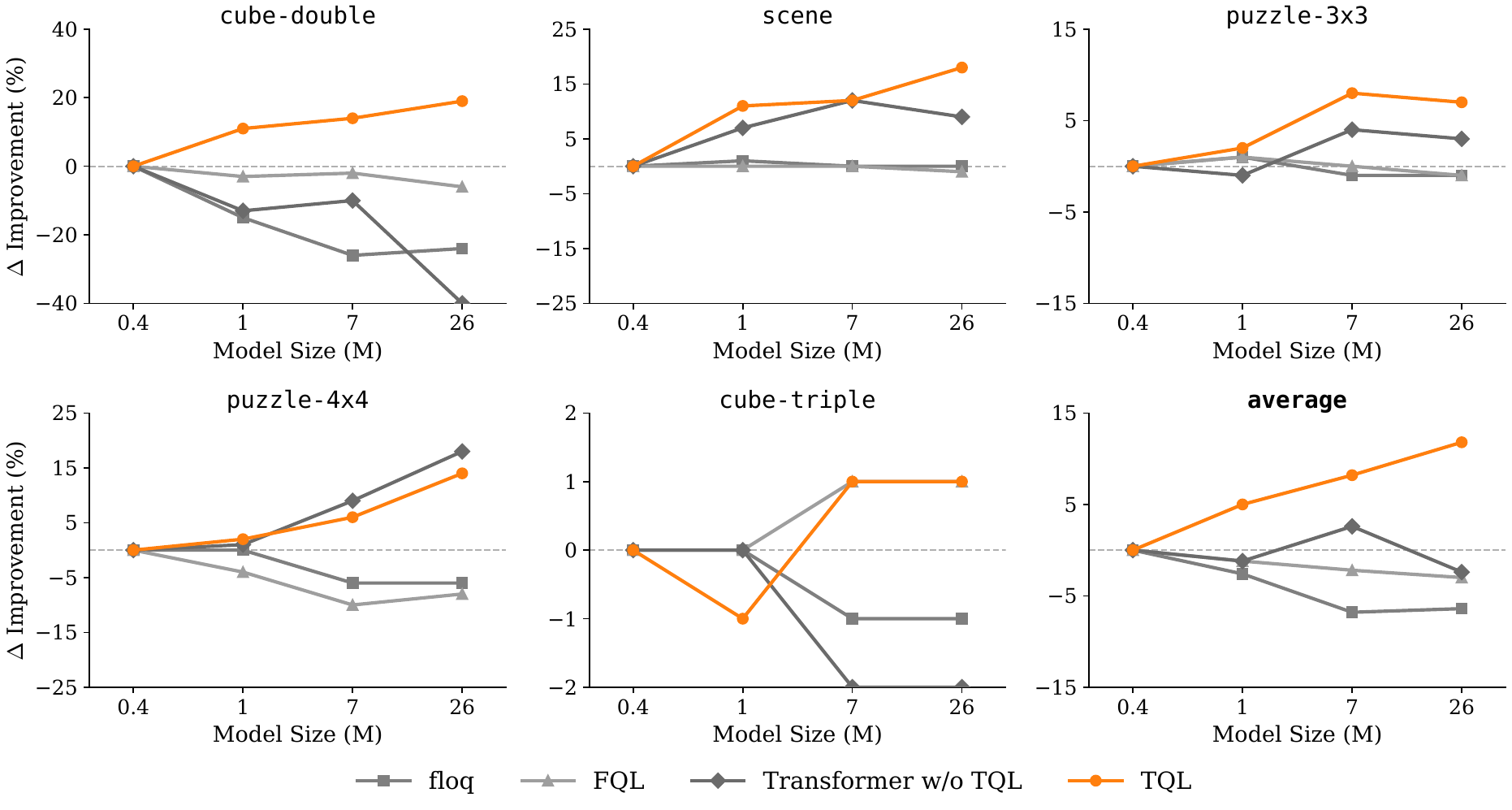}
    \caption{
        \small 
        \textbf{Scaling results.} We compare \ours{} against prior offline RL methods across critic sizes from 0.4M to 26M parameters. The plot reports the average success rate difference compared to the smallest model (0.4M) for each method. While baselines suffer from performance degradation at larger scales, \ours{}  consistently scales well across all environments, outperforming prior methods by a large margin. \looseness=-1
    }
    \label{fig:scaling_exp}
    \vspace{-0.4cm}
\end{figure*}
\subsection{What Prevents Transformers from Scaling Effectively for Value Functions?}\label{sec:scale}
While transformers have demonstrated scalability across numerous domains, their application to value function learning has been limited. We first investigate the specific bottlenecks preventing effective scaling of transformer-based value functions through an empirical analysis. We conduct scaling experiments on the \texttt{cube-double} domain consisting of 5 tasks using the transformer architecture described in \cref{sec:setup}, varying model capacity while maintaining the same architecture. The training curves for different parameter counts are presented in \cref{fig:ana} (left). Contrary to the typical scaling trends observed in language modeling and vision tasks, we observe a strong negative scaling pattern: performance degrades with increased model size, with the largest model performing poorly. 

To diagnose this failure mode, we analyze the Q-value landscapes using a contour visualization in \cref{fig:ana} (center right). We observe that larger networks produce increasingly non-smooth value surfaces and exhibit high-frequency oscillations and discontinuities that are absent in their smaller counterparts. This suggests an instability that occurs with larger networks and that excess capacity appears detrimental. We hypothesize that this degradation is a result of training instabilities in the attention mechanism in transformers. To test this hypothesis, we examine the attention score distributions across network scales in \cref{fig:ana} (right). Consistent with the hypothesis, we find that attention entropy decreases substantially with model size, indicating that larger models learn increasingly peaked and brittle attention patterns. This entropy collapse correlates strongly with both the non-smooth value landscapes and diminished task performance, suggesting that the attention mechanism fails to generalize appropriately when scaled without modifications in the setting of value function learning. We refer to ~\cref{appendix:ae_plots} and \cref{appendix:ae_qmap} for additional attention entropy plots and visualizations for transformer without \ours{} across different environments, which exhibit consistent trends.






\subsection{Scaling \ours{} with Parameter Sizes}
\label{ssec:scaling_exp}
Building upon the empirical analysis in \cref{sec:scale}, we conduct scaling experiments to examine whether \ours{} mitigates attention entropy collapse and enables effective scaling of Q-functions. We compare \ours{} with representative methods for learning the Q-function in the offline RL setting across a variety of generative model backbones, including FQL~\citep{park2025flowqlearning} (MLP), floq~\citep{agrawalla2025floqtrainingcriticsflowmatching} (flow-matching), and PAC~\citep{springenberg2024offlineactorcriticreinforcementlearning} (transformer). We additionally include a transformer baseline as described in \cref{sec:setup}, which has the same architecture as \ours{} but without the attention entropy regularization and modality tokens. All methods are evaluated across a range of critic network sizes ranging from 0.4M to 26M parameters (0.4M, 1M, 7M, and 26M). We note that traditionally, RL methods use network sizes of $\sim$1M parameters to train the Q-function, as evidenced in prior works~\citep{dong2025valueflows,agrawalla2025floqtrainingcriticsflowmatching,kostrikov2021offline,fujimoto2021minimalist,kumar2020conservative} and also by which sizes give the best performance for baselines in our scaling results. \looseness=-1

We present the results in ~\cref{fig:scaling_exp}. From the plots, we see that across all generative model backbones (MLP, flow-matching, transformer), prior methods scale poorly with additional capacity, with an average decrease in performance of 10.6\% from the smallest to largest settings. This suggests that existing offline RL methods, which may have strong performance on offline RL benchmarks, all struggle to benefit from larger value function capacity. Comparing to the transformer baseline, while the performance of the baseline improves on some tasks when scaled up to 7M parameters, it deteriorates at 26M. This behavior is consistent with the attention collapse patterns, as the 26M parameter model shows the most severe collapse and highly non-smooth Q-value landscapes. In contrast to prior methods, \ours{} mitigates this failure mode and achieves stable and consistent scaling, with a 43\% performance improvement from the smallest to the largest model, highlighting the ability of \ours{} to effectively leverage larger capacity for improvement in performance. \looseness=-1




\subsection{Comparing to Prior Offline RL Methods}
Having demonstrated scaling of \ours{}, we further evaluate on a set of benchmark tasks comparing to prior methods to see how well \ours{} can be used to learn a policy from offline datasets. We compare \ours{} against a comprehensive set of offline RL baselines, reporting numbers from the original paper for methods that test on the same benchmarks, or from tuned hyperparameter settings for those that do not. We refer to ~\cref{appendix:baseline} for more details. As summarized in~\cref{tab:offline-eval}, \ours{} achieves the best performance on 4 out of 5 domains, as well as the best average performance across all 25 tasks, demonstrating consistent improvements across a wide range of environments. These results highlight the effectiveness of \ours{} in achieving strong performance in challenging tasks by preventing attention collapse of transformers, resulting in a method that both scales and achieves state-of-the-art performance.

\begin{figure}[t!]
    \centering
    \includegraphics[width=0.9\columnwidth]{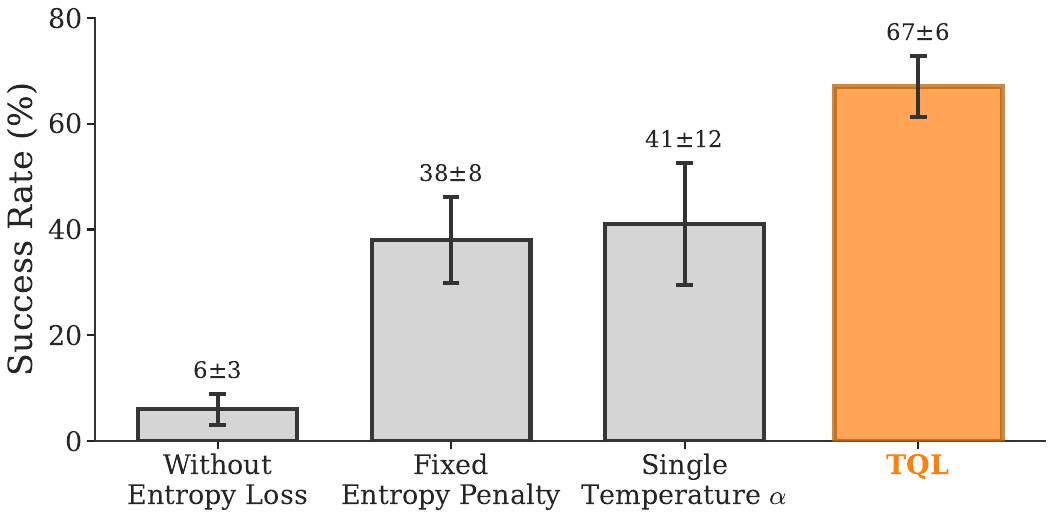}
    \caption{
        \small 
        \textbf{Ablation on the most important components of \ours{}.} We compare \ours{} against four ablated variants on the \texttt{cube-double} environment: (1) a transformer baseline without attention entropy guidance, (2) using a fixed entropy penalty, and (3) using one temperature instead of layer-wise and token-wise temperatures. The results represent the average performance and standard error across 5 tasks and 3 seeds in the \texttt{cube-double}. 
    }
    \vspace{-0.7cm}
    \label{fig:ablation}
\end{figure}

\subsection{Which Components of \ours{} are Most Important?}
\label{sec:ablation}

\begin{figure}[t!]
    \centering
    \includegraphics[width=0.8\columnwidth]{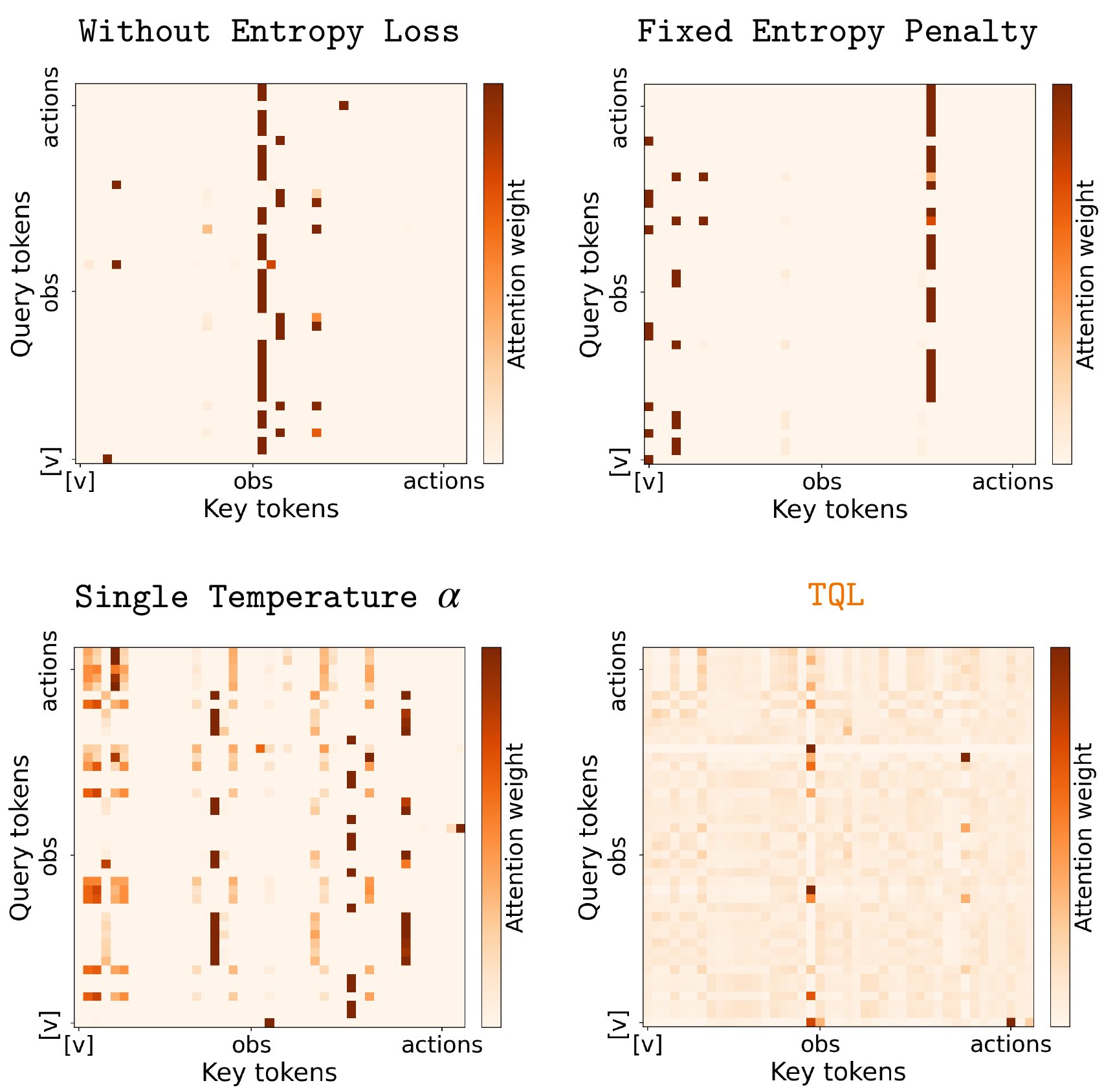}
    \caption{
        \small 
        \textbf{Attention maps of ablations. } We find that \ours{} is the most effective at preventing attention entropy collapse, while both using a fixed entropy penalty and a single temperature across layers and tokens resulted in some degree of over-specialized attention and worse performance. 
    }
    \vspace{-0.6cm}
    \label{fig:vis}
\end{figure}

Having established both scaling and performance, we analyze the key components of \ours{} and how they affect overall performance. We consider the following settings: (1) no entropy loss, which removes the loss term $\mathcal{L}_{\text{attn}}$ in \cref{formula:critic_loss}; (2) fixed entropy penalty, which replaces the entropy adjustment mechanism with a max entropy loss to prevent attention collapse; (3) single temperature $\alpha$, which uses a single temperature across layers and tokens. For each variant, we report performance averaged across 5 tasks in the \texttt{cube-double} environment and using the 26M model. \looseness=-1

The results are shown in ~\cref{fig:ablation}. First, we find that removing entropy guidance entirely leads to a substantial performance drop from attention collapse as we showed in ~\cref{sec:scale}. All forms of entropy adjustment significantly improve performance by preventing attention collapse. We find that using a fixed entropy penalty through a max entropy loss instead of entropy guidance toward a target value results in significantly worse performance, likely because of the instability in the training process arising from the max entropy loss without guiding the entropy toward a target. In addition, layer-wise and token-wise entropy tuning is also important for more stable training, as it is difficult for one temperature term to guide the entropy of all layers toward the same target. We show the attention maps of ablations in ~\cref{fig:vis}. From the attention maps, we see that \ours{} has the most balanced attention across all state and action tokens, which that results in the most effective scaling and highest performance. \looseness=-1
\section{Discussion and Limitations} \label{sec:disc}

We introduce \ours{}, a method for controlling attention entropy which allows for the scaling of transformer-based value functions without collapse. Our work identified that attention entropy collapse is common in transformer-based value functions and found that this collapse has a strong negative correlation with task performance.

While our method demonstrates effective scaling and strong performance on benchmarks, it does have limitations. In terms of implementation complexity, our method adds an additional loss term, as well as an extra hyperparameter for the target entropy. While this adds some complexity and tuning, we found the same set of hyperparameters to work well in our experiments as discussed in \cref{sec:appendix_detail} and, unlike other training stability tricks, our method does not require any architecture changes, enabling the use of pretrained backbones without costly retraining.



\section{Acknowledgments}

This work is in part supported by the RAI Institute, ONR grant N00014-22-1-2293, ONR grant N00014-22-1-2621, NSF \#1941722, and the NSF CAREER award.




\nocite{dong2025reinforcementlearningimplicitimitation,dong2025expostablereinforcementlearning,dong2025mattersbatchonlinereinforcement,wagenmaker2025posteriorbehavioralcloningpretraining}

\bibliography{bibliography}

@misc{kumar2020conservative,
      title={Conservative Q-Learning for Offline Reinforcement Learning}, 
      author={Aviral Kumar and Aurick Zhou and George Tucker and Sergey Levine},
      year={2020},
      eprint={2006.04779},
      archivePrefix={arXiv},
      primaryClass={cs.LG},
      url={https://arxiv.org/abs/2006.04779}, 
}

@misc{intelligence2025pi05visionlanguageactionmodelopenworld,
      title={$\pi_{0.5}$: a Vision-Language-Action Model with Open-World Generalization}, 
      author={{Physical Intelligence} and Kevin Black and Noah Brown and James Darpinian and Karan Dhabalia and Danny Driess and Adnan Esmail and Michael Equi and Chelsea Finn and Niccolo Fusai and Manuel Y. Galliker and Dibya Ghosh and Lachy Groom and Karol Hausman and Brian Ichter and Szymon Jakubczak and Tim Jones and Liyiming Ke and Devin LeBlanc and Sergey Levine and Adrian Li-Bell and Mohith Mothukuri and Suraj Nair and Karl Pertsch and Allen Z. Ren and Lucy Xiaoyang Shi and Laura Smith and Jost Tobias Springenberg and Kyle Stachowicz and James Tanner and Quan Vuong and Homer Walke and Anna Walling and Haohuan Wang and Lili Yu and Ury Zhilinsky},
      year={2025},
      eprint={2504.16054},
      archivePrefix={arXiv},
      primaryClass={cs.LG},
      url={https://arxiv.org/abs/2504.16054}, 
}

@misc{geminiroboticsteam2025geminirobotics15pushing,
      title={Gemini Robotics 1.5: Pushing the Frontier of Generalist Robots with Advanced Embodied Reasoning, Thinking, and Motion Transfer}, 
      author={Gemini Robotics Team and Abbas Abdolmaleki and Saminda Abeyruwan and Joshua Ainslie and Jean-Baptiste Alayrac and Montserrat Gonzalez Arenas and Ashwin Balakrishna and Nathan Batchelor and Alex Bewley and Jeff Bingham and Michael Bloesch and Konstantinos Bousmalis and Philemon Brakel and Anthony Brohan and Thomas Buschmann and Arunkumar Byravan and Serkan Cabi and Ken Caluwaerts and Federico Casarini and Christine Chan and Oscar Chang and London Chappellet-Volpini and Jose Enrique Chen and Xi Chen and Hao-Tien Lewis Chiang and Krzysztof Choromanski and Adrian Collister and David B. D'Ambrosio and Sudeep Dasari and Todor Davchev and Meet Kirankumar Dave and Coline Devin and Norman Di Palo and Tianli Ding and Carl Doersch and Adil Dostmohamed and Yilun Du and Debidatta Dwibedi and Sathish Thoppay Egambaram and Michael Elabd and Tom Erez and Xiaolin Fang and Claudio Fantacci and Cody Fong and Erik Frey and Chuyuan Fu and Ruiqi Gao and Marissa Giustina and Keerthana Gopalakrishnan and Laura Graesser and Oliver Groth and Agrim Gupta and Roland Hafner and Steven Hansen and Leonard Hasenclever and Sam Haves and Nicolas Heess and Brandon Hernaez and Alex Hofer and Jasmine Hsu and Lu Huang and Sandy H. Huang and Atil Iscen and Mithun George Jacob and Deepali Jain and Sally Jesmonth and Abhishek Jindal and Ryan Julian and Dmitry Kalashnikov and M. Emre Karagozler and Stefani Karp and Matija Kecman and J. Chase Kew and Donnie Kim and Frank Kim and Junkyung Kim and Thomas Kipf and Sean Kirmani and Ksenia Konyushkova and Li Yang Ku and Yuheng Kuang and Thomas Lampe and Antoine Laurens and Tuan Anh Le and Isabel Leal and Alex X. Lee and Tsang-Wei Edward Lee and Guy Lever and Jacky Liang and Li-Heng Lin and Fangchen Liu and Shangbang Long and Caden Lu and Sharath Maddineni and Anirudha Majumdar and Kevis-Kokitsi Maninis and Andrew Marmon and Sergio Martinez and Assaf Hurwitz Michaely and Niko Milonopoulos and Joss Moore and Robert Moreno and Michael Neunert and Francesco Nori and Joy Ortiz and Kenneth Oslund and Carolina Parada and Emilio Parisotto and Amaris Paryag and Acorn Pooley and Thomas Power and Alessio Quaglino and Haroon Qureshi and Rajkumar Vasudeva Raju and Helen Ran and Dushyant Rao and Kanishka Rao and Isaac Reid and David Rendleman and Krista Reymann and Miguel Rivas and Francesco Romano and Yulia Rubanova and Peter Pastor Sampedro and Pannag R Sanketi and Dhruv Shah and Mohit Sharma and Kathryn Shea and Mohit Shridhar and Charles Shu and Vikas Sindhwani and Sumeet Singh and Radu Soricut and Rachel Sterneck and Ian Storz and Razvan Surdulescu and Jie Tan and Jonathan Tompson and Saran Tunyasuvunakool and Jake Varley and Grace Vesom and Giulia Vezzani and Maria Bauza Villalonga and Oriol Vinyals and René Wagner and Ayzaan Wahid and Stefan Welker and Paul Wohlhart and Chengda Wu and Markus Wulfmeier and Fei Xia and Ted Xiao and Annie Xie and Jinyu Xie and Peng Xu and Sichun Xu and Ying Xu and Zhuo Xu and Jimmy Yan and Sherry Yang and Skye Yang and Yuxiang Yang and Hiu Hong Yu and Wenhao Yu and Wentao Yuan and Yuan Yuan and Jingwei Zhang and Tingnan Zhang and Zhiyuan Zhang and Allan Zhou and Guangyao Zhou and Yuxiang Zhou},
      year={2025},
      eprint={2510.03342},
      archivePrefix={arXiv},
      primaryClass={cs.RO},
      url={https://arxiv.org/abs/2510.03342}, 
}

@misc{openai2024gpt4ocard,
      title={GPT-4o System Card}, 
      author={OpenAI and : and Aaron Hurst and Adam Lerer and Adam P. Goucher and Adam Perelman and Aditya Ramesh and Aidan Clark and AJ Ostrow and Akila Welihinda and Alan Hayes and Alec Radford and Aleksander Mądry and Alex Baker-Whitcomb and Alex Beutel and Alex Borzunov and Alex Carney and Alex Chow and Alex Kirillov and Alex Nichol and Alex Paino and Alex Renzin and Alex Tachard Passos and Alexander Kirillov and Alexi Christakis and Alexis Conneau and Ali Kamali and Allan Jabri and Allison Moyer and Allison Tam and Amadou Crookes and Amin Tootoochian and Amin Tootoonchian and Ananya Kumar and Andrea Vallone and Andrej Karpathy and Andrew Braunstein and Andrew Cann and Andrew Codispoti and Andrew Galu and Andrew Kondrich and Andrew Tulloch and Andrey Mishchenko and Angela Baek and Angela Jiang and Antoine Pelisse and Antonia Woodford and Anuj Gosalia and Arka Dhar and Ashley Pantuliano and Avi Nayak and Avital Oliver and Barret Zoph and Behrooz Ghorbani and Ben Leimberger and Ben Rossen and Ben Sokolowsky and Ben Wang and Benjamin Zweig and Beth Hoover and Blake Samic and Bob McGrew and Bobby Spero and Bogo Giertler and Bowen Cheng and Brad Lightcap and Brandon Walkin and Brendan Quinn and Brian Guarraci and Brian Hsu and Bright Kellogg and Brydon Eastman and Camillo Lugaresi and Carroll Wainwright and Cary Bassin and Cary Hudson and Casey Chu and Chad Nelson and Chak Li and Chan Jun Shern and Channing Conger and Charlotte Barette and Chelsea Voss and Chen Ding and Cheng Lu and Chong Zhang and Chris Beaumont and Chris Hallacy and Chris Koch and Christian Gibson and Christina Kim and Christine Choi and Christine McLeavey and Christopher Hesse and Claudia Fischer and Clemens Winter and Coley Czarnecki and Colin Jarvis and Colin Wei and Constantin Koumouzelis and Dane Sherburn and Daniel Kappler and Daniel Levin and Daniel Levy and David Carr and David Farhi and David Mely and David Robinson and David Sasaki and Denny Jin and Dev Valladares and Dimitris Tsipras and Doug Li and Duc Phong Nguyen and Duncan Findlay and Edede Oiwoh and Edmund Wong and Ehsan Asdar and Elizabeth Proehl and Elizabeth Yang and Eric Antonow and Eric Kramer and Eric Peterson and Eric Sigler and Eric Wallace and Eugene Brevdo and Evan Mays and Farzad Khorasani and Felipe Petroski Such and Filippo Raso and Francis Zhang and Fred von Lohmann and Freddie Sulit and Gabriel Goh and Gene Oden and Geoff Salmon and Giulio Starace and Greg Brockman and Hadi Salman and Haiming Bao and Haitang Hu and Hannah Wong and Haoyu Wang and Heather Schmidt and Heather Whitney and Heewoo Jun and Hendrik Kirchner and Henrique Ponde de Oliveira Pinto and Hongyu Ren and Huiwen Chang and Hyung Won Chung and Ian Kivlichan and Ian O'Connell and Ian O'Connell and Ian Osband and Ian Silber and Ian Sohl and Ibrahim Okuyucu and Ikai Lan and Ilya Kostrikov and Ilya Sutskever and Ingmar Kanitscheider and Ishaan Gulrajani and Jacob Coxon and Jacob Menick and Jakub Pachocki and James Aung and James Betker and James Crooks and James Lennon and Jamie Kiros and Jan Leike and Jane Park and Jason Kwon and Jason Phang and Jason Teplitz and Jason Wei and Jason Wolfe and Jay Chen and Jeff Harris and Jenia Varavva and Jessica Gan Lee and Jessica Shieh and Ji Lin and Jiahui Yu and Jiayi Weng and Jie Tang and Jieqi Yu and Joanne Jang and Joaquin Quinonero Candela and Joe Beutler and Joe Landers and Joel Parish and Johannes Heidecke and John Schulman and Jonathan Lachman and Jonathan McKay and Jonathan Uesato and Jonathan Ward and Jong Wook Kim and Joost Huizinga and Jordan Sitkin and Jos Kraaijeveld and Josh Gross and Josh Kaplan and Josh Snyder and Joshua Achiam and Joy Jiao and Joyce Lee and Juntang Zhuang and Justyn Harriman and Kai Fricke and Kai Hayashi and Karan Singhal and Katy Shi and Kavin Karthik and Kayla Wood and Kendra Rimbach and Kenny Hsu and Kenny Nguyen and Keren Gu-Lemberg and Kevin Button and Kevin Liu and Kiel Howe and Krithika Muthukumar and Kyle Luther and Lama Ahmad and Larry Kai and Lauren Itow and Lauren Workman and Leher Pathak and Leo Chen and Li Jing and Lia Guy and Liam Fedus and Liang Zhou and Lien Mamitsuka and Lilian Weng and Lindsay McCallum and Lindsey Held and Long Ouyang and Louis Feuvrier and Lu Zhang and Lukas Kondraciuk and Lukasz Kaiser and Luke Hewitt and Luke Metz and Lyric Doshi and Mada Aflak and Maddie Simens and Madelaine Boyd and Madeleine Thompson and Marat Dukhan and Mark Chen and Mark Gray and Mark Hudnall and Marvin Zhang and Marwan Aljubeh and Mateusz Litwin and Matthew Zeng and Max Johnson and Maya Shetty and Mayank Gupta and Meghan Shah and Mehmet Yatbaz and Meng Jia Yang and Mengchao Zhong and Mia Glaese and Mianna Chen and Michael Janner and Michael Lampe and Michael Petrov and Michael Wu and Michele Wang and Michelle Fradin and Michelle Pokrass and Miguel Castro and Miguel Oom Temudo de Castro and Mikhail Pavlov and Miles Brundage and Miles Wang and Minal Khan and Mira Murati and Mo Bavarian and Molly Lin and Murat Yesildal and Nacho Soto and Natalia Gimelshein and Natalie Cone and Natalie Staudacher and Natalie Summers and Natan LaFontaine and Neil Chowdhury and Nick Ryder and Nick Stathas and Nick Turley and Nik Tezak and Niko Felix and Nithanth Kudige and Nitish Keskar and Noah Deutsch and Noel Bundick and Nora Puckett and Ofir Nachum and Ola Okelola and Oleg Boiko and Oleg Murk and Oliver Jaffe and Olivia Watkins and Olivier Godement and Owen Campbell-Moore and Patrick Chao and Paul McMillan and Pavel Belov and Peng Su and Peter Bak and Peter Bakkum and Peter Deng and Peter Dolan and Peter Hoeschele and Peter Welinder and Phil Tillet and Philip Pronin and Philippe Tillet and Prafulla Dhariwal and Qiming Yuan and Rachel Dias and Rachel Lim and Rahul Arora and Rajan Troll and Randall Lin and Rapha Gontijo Lopes and Raul Puri and Reah Miyara and Reimar Leike and Renaud Gaubert and Reza Zamani and Ricky Wang and Rob Donnelly and Rob Honsby and Rocky Smith and Rohan Sahai and Rohit Ramchandani and Romain Huet and Rory Carmichael and Rowan Zellers and Roy Chen and Ruby Chen and Ruslan Nigmatullin and Ryan Cheu and Saachi Jain and Sam Altman and Sam Schoenholz and Sam Toizer and Samuel Miserendino and Sandhini Agarwal and Sara Culver and Scott Ethersmith and Scott Gray and Sean Grove and Sean Metzger and Shamez Hermani and Shantanu Jain and Shengjia Zhao and Sherwin Wu and Shino Jomoto and Shirong Wu and Shuaiqi and Xia and Sonia Phene and Spencer Papay and Srinivas Narayanan and Steve Coffey and Steve Lee and Stewart Hall and Suchir Balaji and Tal Broda and Tal Stramer and Tao Xu and Tarun Gogineni and Taya Christianson and Ted Sanders and Tejal Patwardhan and Thomas Cunninghman and Thomas Degry and Thomas Dimson and Thomas Raoux and Thomas Shadwell and Tianhao Zheng and Todd Underwood and Todor Markov and Toki Sherbakov and Tom Rubin and Tom Stasi and Tomer Kaftan and Tristan Heywood and Troy Peterson and Tyce Walters and Tyna Eloundou and Valerie Qi and Veit Moeller and Vinnie Monaco and Vishal Kuo and Vlad Fomenko and Wayne Chang and Weiyi Zheng and Wenda Zhou and Wesam Manassra and Will Sheu and Wojciech Zaremba and Yash Patil and Yilei Qian and Yongjik Kim and Youlong Cheng and Yu Zhang and Yuchen He and Yuchen Zhang and Yujia Jin and Yunxing Dai and Yury Malkov},
      year={2024},
      eprint={2410.21276},
      archivePrefix={arXiv},
      primaryClass={cs.CL},
      url={https://arxiv.org/abs/2410.21276}, 
}

@misc{siméoni2025dinov3,
      title={DINOv3}, 
      author={Oriane Siméoni and Huy V. Vo and Maximilian Seitzer and Federico Baldassarre and Maxime Oquab and Cijo Jose and Vasil Khalidov and Marc Szafraniec and Seungeun Yi and Michaël Ramamonjisoa and Francisco Massa and Daniel Haziza and Luca Wehrstedt and Jianyuan Wang and Timothée Darcet and Théo Moutakanni and Leonel Sentana and Claire Roberts and Andrea Vedaldi and Jamie Tolan and John Brandt and Camille Couprie and Julien Mairal and Hervé Jégou and Patrick Labatut and Piotr Bojanowski},
      year={2025},
      eprint={2508.10104},
      archivePrefix={arXiv},
      primaryClass={cs.CV},
      url={https://arxiv.org/abs/2508.10104}, 
}

@misc{yu2020mopo,
      title={MOPO: Model-based Offline Policy Optimization}, 
      author={Tianhe Yu and Garrett Thomas and Lantao Yu and Stefano Ermon and James Zou and Sergey Levine and Chelsea Finn and Tengyu Ma},
      year={2020},
      eprint={2005.13239},
      archivePrefix={arXiv},
      primaryClass={cs.LG},
      url={https://arxiv.org/abs/2005.13239}, 
}

@misc{wu2021uncertainty,
      title={Uncertainty Weighted Actor-Critic for Offline Reinforcement Learning}, 
      author={Yue Wu and Shuangfei Zhai and Nitish Srivastava and Joshua Susskind and Jian Zhang and Ruslan Salakhutdinov and Hanlin Goh},
      year={2021},
      eprint={2105.08140},
      archivePrefix={arXiv},
      primaryClass={cs.LG},
      url={https://arxiv.org/abs/2105.08140}, 
}

@misc{fujimoto2019off,
      title={Off-Policy Deep Reinforcement Learning without Exploration}, 
      author={Scott Fujimoto and David Meger and Doina Precup},
      year={2019},
      eprint={1812.02900},
      archivePrefix={arXiv},
      primaryClass={cs.LG},
      url={https://arxiv.org/abs/1812.02900}, 
}

@misc{kostrikov2021offline,
      title={Offline Reinforcement Learning with Implicit Q-Learning}, 
      author={Ilya Kostrikov and Ashvin Nair and Sergey Levine},
      year={2021},
      eprint={2110.06169},
      archivePrefix={arXiv},
      primaryClass={cs.LG},
      url={https://arxiv.org/abs/2110.06169}, 
}

@misc{fujimoto2021minimalist,
      title={A Minimalist Approach to Offline Reinforcement Learning}, 
      author={Scott Fujimoto and Shixiang Shane Gu},
      year={2021},
      eprint={2106.06860},
      archivePrefix={arXiv},
      primaryClass={cs.LG},
      url={https://arxiv.org/abs/2106.06860}, 
}

@misc{nair2021awac,
      title={AWAC: Accelerating Online Reinforcement Learning with Offline Datasets}, 
      author={Ashvin Nair and Abhishek Gupta and Murtaza Dalal and Sergey Levine},
      year={2021},
      eprint={2006.09359},
      archivePrefix={arXiv},
      primaryClass={cs.LG},
      url={https://arxiv.org/abs/2006.09359}, 
}

@misc{kidambi2020morel,
      title={MOReL : Model-Based Offline Reinforcement Learning}, 
      author={Rahul Kidambi and Aravind Rajeswaran and Praneeth Netrapalli and Thorsten Joachims},
      year={2021},
      eprint={2005.05951},
      archivePrefix={arXiv},
      primaryClass={cs.LG},
      url={https://arxiv.org/abs/2005.05951}, 
}

@misc{hansenestruch2023idqlimplicitqlearningactorcritic,
      title={IDQL: Implicit Q-Learning as an Actor-Critic Method with Diffusion Policies}, 
      author={Philippe Hansen-Estruch and Ilya Kostrikov and Michael Janner and Jakub Grudzien Kuba and Sergey Levine},
      year={2023},
      eprint={2304.10573},
      archivePrefix={arXiv},
      primaryClass={cs.LG},
      url={https://arxiv.org/abs/2304.10573}, 
}

@misc{wu2019behaviorregularizedofflinereinforcement,
      title={Behavior Regularized Offline Reinforcement Learning}, 
      author={Yifan Wu and George Tucker and Ofir Nachum},
      year={2019},
      eprint={1911.11361},
      archivePrefix={arXiv},
      primaryClass={cs.LG},
      url={https://arxiv.org/abs/1911.11361}, 
}

@misc{castanyer2025stablegradientsstablelearning,
      title={Stable Gradients for Stable Learning at Scale in Deep Reinforcement Learning}, 
      author={Roger Creus Castanyer and Johan Obando-Ceron and Lu Li and Pierre-Luc Bacon and Glen Berseth and Aaron Courville and Pablo Samuel Castro},
      year={2025},
      eprint={2506.15544},
      archivePrefix={arXiv},
      primaryClass={cs.LG},
      url={https://arxiv.org/abs/2506.15544}, 
}

@misc{tarasov2023revisitingminimalistapproachoffline,
      title={Revisiting the Minimalist Approach to Offline Reinforcement Learning}, 
      author={Denis Tarasov and Vladislav Kurenkov and Alexander Nikulin and Sergey Kolesnikov},
      year={2023},
      eprint={2305.09836},
      archivePrefix={arXiv},
      primaryClass={cs.LG},
      url={https://arxiv.org/abs/2305.09836}, 
}

@article{park2025flowqlearning,
  title={Flow q-learning},
  author={Park, Seohong and Li, Qiyang and Levine, Sergey},
  journal={arXiv preprint arXiv:2502.02538},
  year={2025}
}

@misc{chen2021decision,
      title={Decision Transformer: Reinforcement Learning via Sequence Modeling}, 
      author={Lili Chen and Kevin Lu and Aravind Rajeswaran and Kimin Lee and Aditya Grover and Michael Laskin and Pieter Abbeel and Aravind Srinivas and Igor Mordatch},
      year={2021},
      eprint={2106.01345},
      archivePrefix={arXiv},
      primaryClass={cs.LG},
      url={https://arxiv.org/abs/2106.01345}, 
}

@misc{furuta2022generalizeddecisiontransformeroffline,
      title={Generalized Decision Transformer for Offline Hindsight Information Matching}, 
      author={Hiroki Furuta and Yutaka Matsuo and Shixiang Shane Gu},
      year={2022},
      eprint={2111.10364},
      archivePrefix={arXiv},
      primaryClass={cs.LG},
      url={https://arxiv.org/abs/2111.10364}, 
}

@misc{janner2021offlinereinforcementlearningbig,
      title={Offline Reinforcement Learning as One Big Sequence Modeling Problem}, 
      author={Michael Janner and Qiyang Li and Sergey Levine},
      year={2021},
      eprint={2106.02039},
      archivePrefix={arXiv},
      primaryClass={cs.LG},
      url={https://arxiv.org/abs/2106.02039}, 
}

@misc{zheng2022onlinedecisiontransformer,
      title={Online Decision Transformer}, 
      author={Qinqing Zheng and Amy Zhang and Aditya Grover},
      year={2022},
      eprint={2202.05607},
      archivePrefix={arXiv},
      primaryClass={cs.LG},
      url={https://arxiv.org/abs/2202.05607}, 
}

@misc{wu2023elasticdecisiontransformer,
      title={Elastic Decision Transformer}, 
      author={Yueh-Hua Wu and Xiaolong Wang and Masashi Hamaya},
      year={2023},
      eprint={2307.02484},
      archivePrefix={arXiv},
      primaryClass={cs.LG},
      url={https://arxiv.org/abs/2307.02484}, 
}

@misc{springenberg2024offlineactorcriticreinforcementlearning,
      title={Offline Actor-Critic Reinforcement Learning Scales to Large Models}, 
      author={Jost Tobias Springenberg and Abbas Abdolmaleki and Jingwei Zhang and Oliver Groth and Michael Bloesch and Thomas Lampe and Philemon Brakel and Sarah Bechtle and Steven Kapturowski and Roland Hafner and Nicolas Heess and Martin Riedmiller},
      year={2024},
      eprint={2402.05546},
      archivePrefix={arXiv},
      primaryClass={cs.LG},
      url={https://arxiv.org/abs/2402.05546}, 
}

@misc{agrawalla2025floqtrainingcriticsflowmatching,
      title={floq: Training Critics via Flow-Matching for Scaling Compute in Value-Based RL}, 
      author={Bhavya Agrawalla and Michal Nauman and Khush Agrawal and Aviral Kumar},
      year={2025},
      eprint={2509.06863},
      archivePrefix={arXiv},
      primaryClass={cs.LG},
      url={https://arxiv.org/abs/2509.06863}, 
}

@misc{dong2025valueflows,
      title={Value Flows}, 
      author={Perry Dong and Chongyi Zheng and Chelsea Finn and Dorsa Sadigh and Benjamin Eysenbach},
      year={2025},
      eprint={2510.07650},
      archivePrefix={arXiv},
      primaryClass={cs.LG},
      url={https://arxiv.org/abs/2510.07650}, 
}

@misc{intelligence2025pi06vlalearnsexperience,
      title={$\pi^{*}_{0.6}$: a VLA That Learns From Experience}, 
      author={{Physical Intelligence} and Ali Amin and Raichelle Aniceto and Ashwin Balakrishna and Kevin Black and Ken Conley and Grace Connors and James Darpinian and Karan Dhabalia and Jared DiCarlo and Danny Driess and Michael Equi and Adnan Esmail and Yunhao Fang and Chelsea Finn and Catherine Glossop and Thomas Godden and Ivan Goryachev and Lachy Groom and Hunter Hancock and Karol Hausman and Gashon Hussein and Brian Ichter and Szymon Jakubczak and Rowan Jen and Tim Jones and Ben Katz and Liyiming Ke and Chandra Kuchi and Marinda Lamb and Devin LeBlanc and Sergey Levine and Adrian Li-Bell and Yao Lu and Vishnu Mano and Mohith Mothukuri and Suraj Nair and Karl Pertsch and Allen Z. Ren and Charvi Sharma and Lucy Xiaoyang Shi and Laura Smith and Jost Tobias Springenberg and Kyle Stachowicz and Will Stoeckle and Alex Swerdlow and James Tanner and Marcel Torne and Quan Vuong and Anna Walling and Haohuan Wang and Blake Williams and Sukwon Yoo and Lili Yu and Ury Zhilinsky and Zhiyuan Zhou},
      year={2025},
      eprint={2511.14759},
      archivePrefix={arXiv},
      primaryClass={cs.LG},
      url={https://arxiv.org/abs/2511.14759}, 
}

@misc{chebotar2023qtransformer,
      title={Q-Transformer: Scalable Offline Reinforcement Learning via Autoregressive Q-Functions}, 
      author={Yevgen Chebotar and Quan Vuong and Alex Irpan and Karol Hausman and Fei Xia and Yao Lu and Aviral Kumar and Tianhe Yu and Alexander Herzog and Karl Pertsch and Keerthana Gopalakrishnan and Julian Ibarz and Ofir Nachum and Sumedh Sontakke and Grecia Salazar and Huong T Tran and Jodilyn Peralta and Clayton Tan and Deeksha Manjunath and Jaspiar Singht and Brianna Zitkovich and Tomas Jackson and Kanishka Rao and Chelsea Finn and Sergey Levine},
      year={2023},
      eprint={2309.10150},
      archivePrefix={arXiv},
      primaryClass={cs.RO},
      url={https://arxiv.org/abs/2309.10150}, 
}

@misc{obandoceron2024mixturesexpertsunlockparameter,
      title={Mixtures of Experts Unlock Parameter Scaling for Deep RL}, 
      author={Johan Obando-Ceron and Ghada Sokar and Timon Willi and Clare Lyle and Jesse Farebrother and Jakob Foerster and Gintare Karolina Dziugaite and Doina Precup and Pablo Samuel Castro},
      year={2024},
      eprint={2402.08609},
      archivePrefix={arXiv},
      primaryClass={cs.LG},
      url={https://arxiv.org/abs/2402.08609}, 
}

@misc{fan2024scaling,
      title={Scaling Offline Model-Based RL via Jointly-Optimized World-Action Model Pretraining}, 
      author={Jie Cheng and Ruixi Qiao and Yingwei Ma and Binhua Li and Gang Xiong and Qinghai Miao and Yongbin Li and Yisheng Lv},
      year={2025},
      eprint={2410.00564},
      archivePrefix={arXiv},
      primaryClass={cs.LG},
      url={https://arxiv.org/abs/2410.00564}, 
}

@misc{yoshida2017spectralnormregularizationimproving,
      title={Spectral Norm Regularization for Improving the Generalizability of Deep Learning}, 
      author={Yuichi Yoshida and Takeru Miyato},
      year={2017},
      eprint={1705.10941},
      archivePrefix={arXiv},
      primaryClass={stat.ML},
      url={https://arxiv.org/abs/1705.10941}, 
}

@misc{shazeer2020gluvariantsimprovetransformer,
      title={GLU Variants Improve Transformer}, 
      author={Noam Shazeer},
      year={2020},
      eprint={2002.05202},
      archivePrefix={arXiv},
      primaryClass={cs.LG},
      url={https://arxiv.org/abs/2002.05202}, 
}

@misc{ding2021cogviewmasteringtexttoimagegeneration,
      title={CogView: Mastering Text-to-Image Generation via Transformers}, 
      author={Ming Ding and Zhuoyi Yang and Wenyi Hong and Wendi Zheng and Chang Zhou and Da Yin and Junyang Lin and Xu Zou and Zhou Shao and Hongxia Yang and Jie Tang},
      year={2021},
      eprint={2105.13290},
      archivePrefix={arXiv},
      primaryClass={cs.CV},
      url={https://arxiv.org/abs/2105.13290}, 
}

@misc{zhang2019rootmeansquarelayer,
      title={Root Mean Square Layer Normalization}, 
      author={Biao Zhang and Rico Sennrich},
      year={2019},
      eprint={1910.07467},
      archivePrefix={arXiv},
      primaryClass={cs.LG},
      url={https://arxiv.org/abs/1910.07467}, 
}

@misc{zhai2023stabilizing,
      title={Stabilizing Transformer Training by Preventing Attention Entropy Collapse}, 
      author={Shuangfei Zhai and Tatiana Likhomanenko and Etai Littwin and Dan Busbridge and Jason Ramapuram and Yizhe Zhang and Jiatao Gu and Josh Susskind},
      year={2023},
      eprint={2303.06296},
      archivePrefix={arXiv},
      primaryClass={cs.LG},
      url={https://arxiv.org/abs/2303.06296}, 
}

@misc{wortsman2023smallscale,
      title={Small-scale proxies for large-scale Transformer training instabilities}, 
      author={Mitchell Wortsman and Peter J. Liu and Lechao Xiao and Katie Everett and Alex Alemi and Ben Adlam and John D. Co-Reyes and Izzeddin Gur and Abhishek Kumar and Roman Novak and Jeffrey Pennington and Jascha Sohl-dickstein and Kelvin Xu and Jaehoon Lee and Justin Gilmer and Simon Kornblith},
      year={2023},
      eprint={2309.14322},
      archivePrefix={arXiv},
      primaryClass={cs.LG},
      url={https://arxiv.org/abs/2309.14322}, 
}

@misc{lai2024hybridnorm,
      title={HybridNorm: Towards Stable and Efficient Transformer Training via Hybrid Normalization}, 
      author={Zhijian Zhuo and Yutao Zeng and Ya Wang and Sijun Zhang and Jian Yang and Xiaoqing Li and Xun Zhou and Jinwen Ma},
      year={2025},
      eprint={2503.04598},
      archivePrefix={arXiv},
      primaryClass={cs.CL},
      url={https://arxiv.org/abs/2503.04598}, 
}

@misc{schwarzer2023bigger,
      title={Bigger, Better, Faster: Human-level Atari with human-level efficiency}, 
      author={Max Schwarzer and Johan Obando-Ceron and Aaron Courville and Marc Bellemare and Rishabh Agarwal and Pablo Samuel Castro},
      year={2023},
      eprint={2305.19452},
      archivePrefix={arXiv},
      primaryClass={cs.LG},
      url={https://arxiv.org/abs/2305.19452}, 
}

@misc{nauman2024biggerregularizedoptimisticscaling,
      title={Bigger, Regularized, Optimistic: scaling for compute and sample-efficient continuous control}, 
      author={Michal Nauman and Mateusz Ostaszewski and Krzysztof Jankowski and Piotr Miłoś and Marek Cygan},
      year={2024},
      eprint={2405.16158},
      archivePrefix={arXiv},
      primaryClass={cs.LG},
      url={https://arxiv.org/abs/2405.16158}, 
}

@misc{farebrother2024stop,
      title={Stop Regressing: Training Value Functions via Classification for Scalable Deep RL}, 
      author={Jesse Farebrother and Jordi Orbay and Quan Vuong and Adrien Ali Taïga and Yevgen Chebotar and Ted Xiao and Alex Irpan and Sergey Levine and Pablo Samuel Castro and Aleksandra Faust and Aviral Kumar and Rishabh Agarwal},
      year={2024},
      eprint={2403.03950},
      archivePrefix={arXiv},
      primaryClass={cs.LG},
      url={https://arxiv.org/abs/2403.03950}, 
}

@misc{wang20251000layernetworksselfsupervised,
      title={1000 Layer Networks for Self-Supervised RL: Scaling Depth Can Enable New Goal-Reaching Capabilities}, 
      author={Kevin Wang and Ishaan Javali and Michał Bortkiewicz and Tomasz Trzciński and Benjamin Eysenbach},
      year={2025},
      eprint={2503.14858},
      archivePrefix={arXiv},
      primaryClass={cs.LG},
      url={https://arxiv.org/abs/2503.14858}, 
}

@misc{haarnoja2018softactorcriticoffpolicymaximum,
      title={Soft Actor-Critic: Off-Policy Maximum Entropy Deep Reinforcement Learning with a Stochastic Actor}, 
      author={Tuomas Haarnoja and Aurick Zhou and Pieter Abbeel and Sergey Levine},
      year={2018},
      eprint={1801.01290},
      archivePrefix={arXiv},
      primaryClass={cs.LG},
      url={https://arxiv.org/abs/1801.01290}, 
}

@misc{dong2025reinforcementlearningimplicitimitation,
      title={Reinforcement Learning via Implicit Imitation Guidance}, 
      author={Perry Dong and Alec M. Lessing and Annie S. Chen and Chelsea Finn},
      year={2025},
      eprint={2506.07505},
      archivePrefix={arXiv},
      primaryClass={cs.LG},
      url={https://arxiv.org/abs/2506.07505}, 
}

@misc{dong2025expostablereinforcementlearning,
      title={EXPO: Stable Reinforcement Learning with Expressive Policies}, 
      author={Perry Dong and Qiyang Li and Dorsa Sadigh and Chelsea Finn},
      year={2025},
      eprint={2507.07986},
      archivePrefix={arXiv},
      primaryClass={cs.LG},
      url={https://arxiv.org/abs/2507.07986}, 
}

@misc{dong2025mattersbatchonlinereinforcement,
      title={What Matters for Batch Online Reinforcement Learning in Robotics?}, 
      author={Perry Dong and Suvir Mirchandani and Dorsa Sadigh and Chelsea Finn},
      year={2025},
      eprint={2505.08078},
      archivePrefix={arXiv},
      primaryClass={cs.RO},
      url={https://arxiv.org/abs/2505.08078}, 
}

@misc{wagenmaker2025posteriorbehavioralcloningpretraining,
      title={Posterior Behavioral Cloning: Pretraining BC Policies for Efficient RL Finetuning}, 
      author={Andrew Wagenmaker and Perry Dong and Raymond Tsao and Chelsea Finn and Sergey Levine},
      year={2025},
      eprint={2512.16911},
      archivePrefix={arXiv},
      primaryClass={cs.LG},
      url={https://arxiv.org/abs/2512.16911}, 
}

@misc{brohan2023rt1roboticstransformerrealworld,
      title={RT-1: Robotics Transformer for Real-World Control at Scale}, 
      author={Anthony Brohan and Noah Brown and Justice Carbajal and Yevgen Chebotar and Joseph Dabis and Chelsea Finn and Keerthana Gopalakrishnan and Karol Hausman and Alex Herzog and Jasmine Hsu and Julian Ibarz and Brian Ichter and Alex Irpan and Tomas Jackson and Sally Jesmonth and Nikhil J Joshi and Ryan Julian and Dmitry Kalashnikov and Yuheng Kuang and Isabel Leal and Kuang-Huei Lee and Sergey Levine and Yao Lu and Utsav Malla and Deeksha Manjunath and Igor Mordatch and Ofir Nachum and Carolina Parada and Jodilyn Peralta and Emily Perez and Karl Pertsch and Jornell Quiambao and Kanishka Rao and Michael Ryoo and Grecia Salazar and Pannag Sanketi and Kevin Sayed and Jaspiar Singh and Sumedh Sontakke and Austin Stone and Clayton Tan and Huong Tran and Vincent Vanhoucke and Steve Vega and Quan Vuong and Fei Xia and Ted Xiao and Peng Xu and Sichun Xu and Tianhe Yu and Brianna Zitkovich},
      year={2023},
      eprint={2212.06817},
      archivePrefix={arXiv},
      primaryClass={cs.RO},
      url={https://arxiv.org/abs/2212.06817}, 
}

@misc{rybakov2024methodsimprovingllmtraining,
      title={Methods of improving LLM training stability}, 
      author={Oleg Rybakov and Mike Chrzanowski and Peter Dykas and Jinze Xue and Ben Lanir},
      year={2024},
      eprint={2410.16682},
      archivePrefix={arXiv},
      primaryClass={cs.CL},
      url={https://arxiv.org/abs/2410.16682}, 
}

@misc{park2025ogbench,
      title={OGBench: Benchmarking Offline Goal-Conditioned RL}, 
      author={Seohong Park and Kevin Frans and Benjamin Eysenbach and Sergey Levine},
      year={2025},
      eprint={2410.20092},
      archivePrefix={arXiv},
      primaryClass={cs.LG},
      url={https://arxiv.org/abs/2410.20092}, 
}

@misc{lee2025simbasimplicitybiasscaling,
      title={SimBa: Simplicity Bias for Scaling Up Parameters in Deep Reinforcement Learning}, 
      author={Hojoon Lee and Dongyoon Hwang and Donghu Kim and Hyunseung Kim and Jun Jet Tai and Kaushik Subramanian and Peter R. Wurman and Jaegul Choo and Peter Stone and Takuma Seno},
      year={2025},
      eprint={2410.09754},
      archivePrefix={arXiv},
      primaryClass={cs.LG},
      url={https://arxiv.org/abs/2410.09754}, 
}

@misc{levine2020offlinereinforcementlearningtutorial,
      title={Offline Reinforcement Learning: Tutorial, Review, and Perspectives on Open Problems}, 
      author={Sergey Levine and Aviral Kumar and George Tucker and Justin Fu},
      year={2020},
      eprint={2005.01643},
      archivePrefix={arXiv},
      primaryClass={cs.LG},
      url={https://arxiv.org/abs/2005.01643}, 
}

@inproceedings{dehghani2023scaling,
  title={Scaling vision transformers to 22 billion parameters},
  author={Dehghani, Mostafa and Djolonga, Josip and Mustafa, Basil and Padlewski, Piotr and Heek, Jonathan and Gilmer, Justin and Steiner, Andreas Peter and Caron, Mathilde and Geirhos, Robert and Alabdulmohsin, Ibrahim and others},
  booktitle={International conference on machine learning},
  pages={7480--7512},
  year={2023},
  organization={PMLR}
}
\bibliographystyle{icml2026}


\newpage
\appendix
\onecolumn
\section{Full Scaling Analysis}


\subsection{Full Attention Entropy Plots for Transformer Without \ours{}} 
\label{appendix:ae_plots}
\begin{figure*}[h!]
    \centering
    \includegraphics[width=0.99\linewidth]{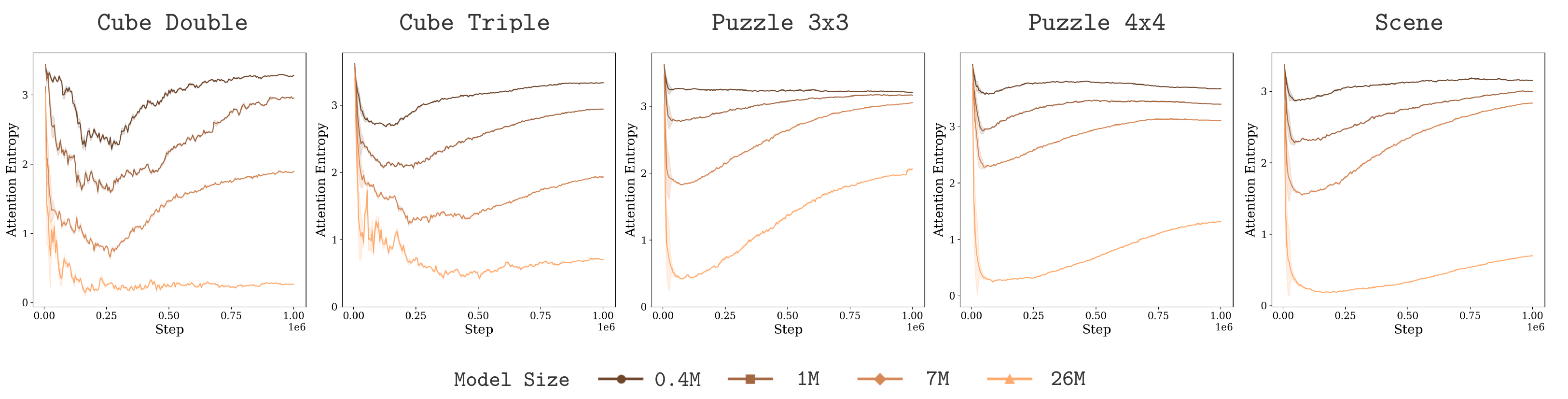}
    \caption{\textbf{Attention entropy analysis for Transformer without \ours{}.} We visualize attention entropy across 5 environments as model size increases. The plots demonstrate a consistent trend of decreasing entropy with larger models, providing empirical evidence of the attention collapse phenomenon.} 
    \label{fig:ae_plots}
\end{figure*}

In this section, we examine the attention entropy of a Transformer without \ours{} across all five environments, as shown in \Cref{fig:ae_plots}. We observe a consistent trend where attention entropy decreases as model size increases, eventually collapsing to near-zero values for the largest models. This empirical evidence validates the attention collapse problem identified in our analysis.

Furthermore, we find that these entropy profiles are highly predictive of the scaling behavior reported in Figure~\ref{fig:scaling_exp}. Specifically, the \texttt{puzzle-3x3} and \texttt{puzzle-4x4} environments exhibit the least severe entropy collapse in Figure~\ref{fig:ae_plots}. Correspondingly, the Transformer without \ours{} is able to achieve slight positive scaling on these same tasks in Figure~\ref{fig:scaling_exp}. This strong correlation between sustained attention entropy and scaling ability reinforces our hypothesis that attention collapse is a primary bottleneck for performance at scale.

\subsection{Attention Map and Q-Value Visualization} \label{appendix:ae_qmap}
In this section, we present comprehensive visualizations comparing the attention maps and Q-value landscapes. We evaluate \ours{} with 26M parameters against a Transformer without \ours{} baseline trained without attention entropy tuning. To ensure a fair comparison, we utilize identical state and action pairs drawn from random successful trajectories across five environments.

\begin{figure*}[h!]
    \centering
    \includegraphics[width=0.99\linewidth]{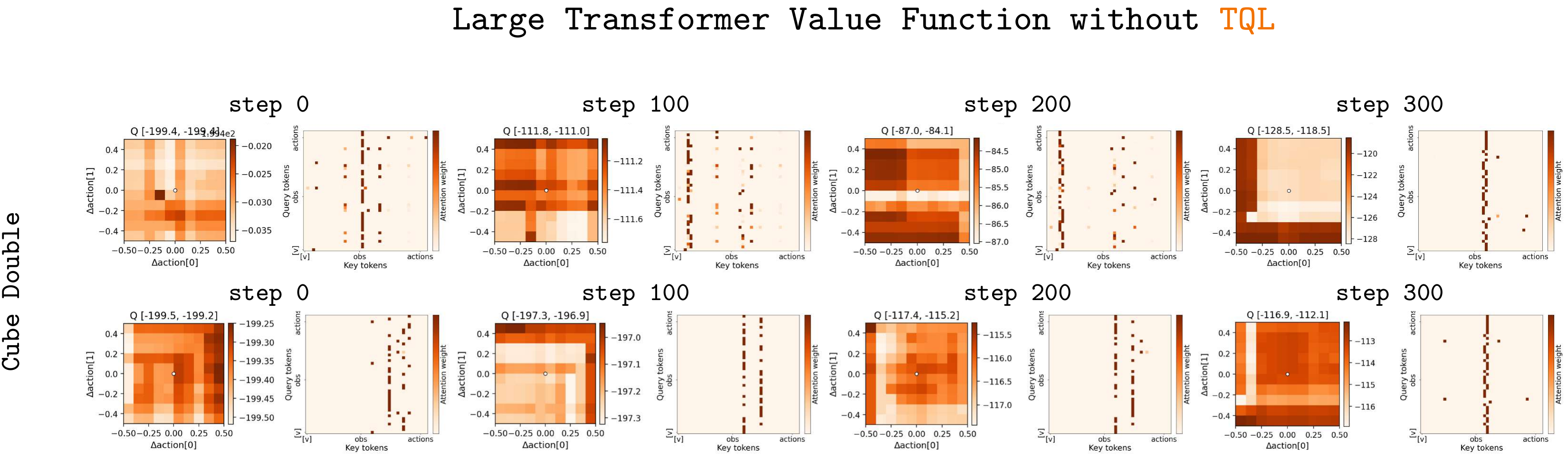}
    \vspace{1pt}
    \includegraphics[width=0.99\linewidth]{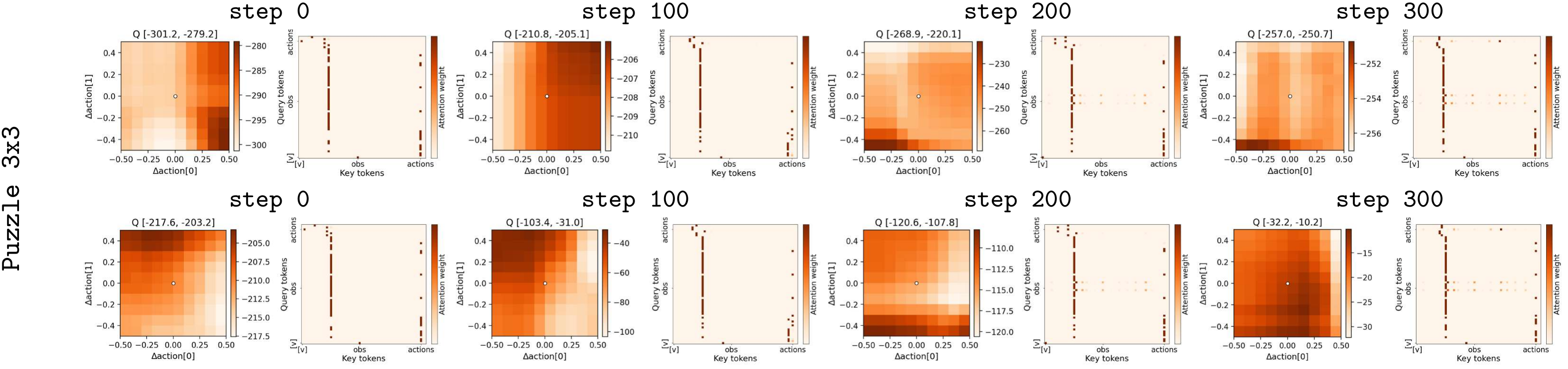}
    \vspace{1pt}
    \includegraphics[width=0.99\linewidth]{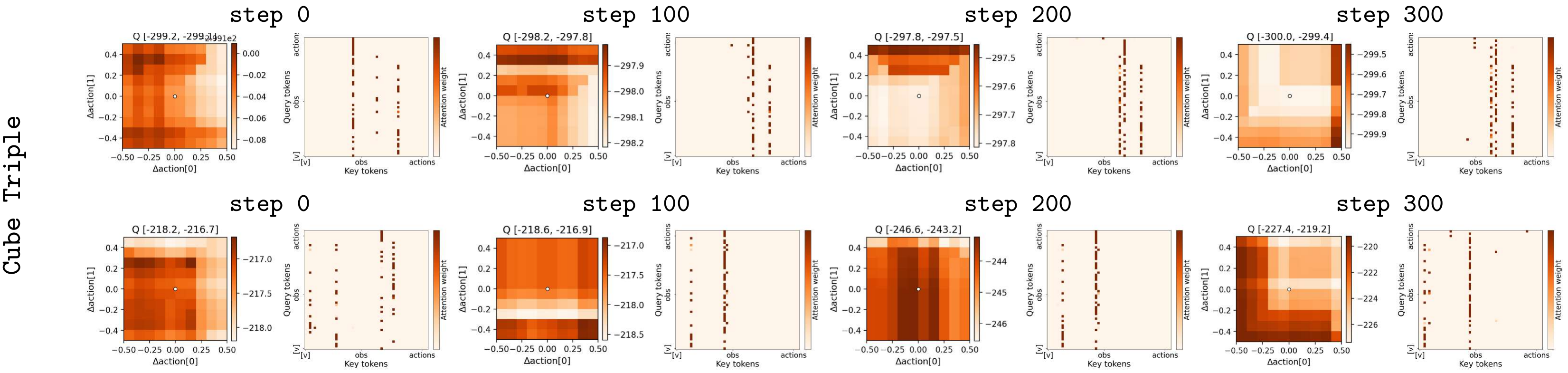}
    \vspace{1pt}
    \includegraphics[width=0.99\linewidth]{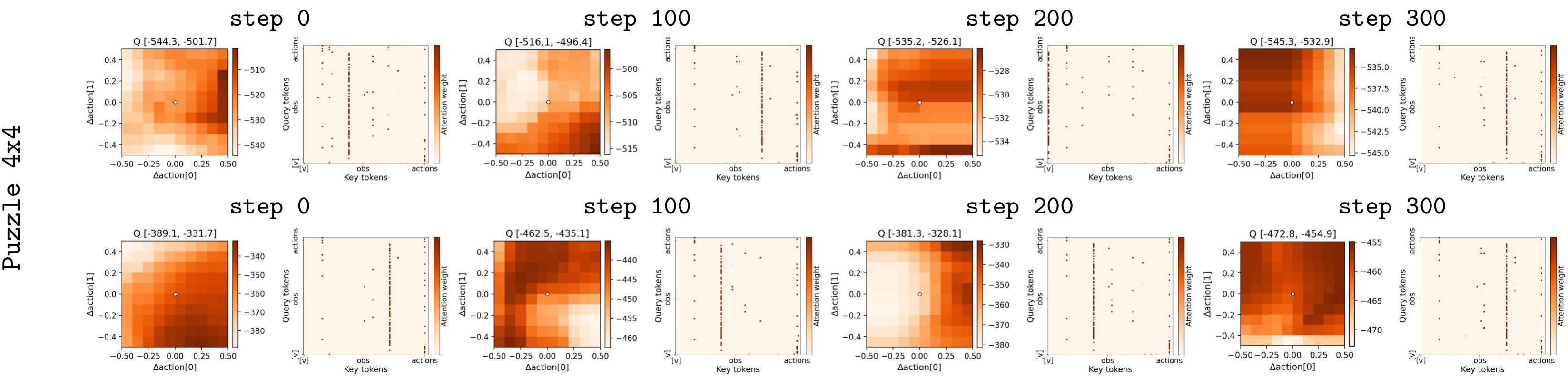}
    \vspace{1pt}
    \includegraphics[width=0.99\linewidth]{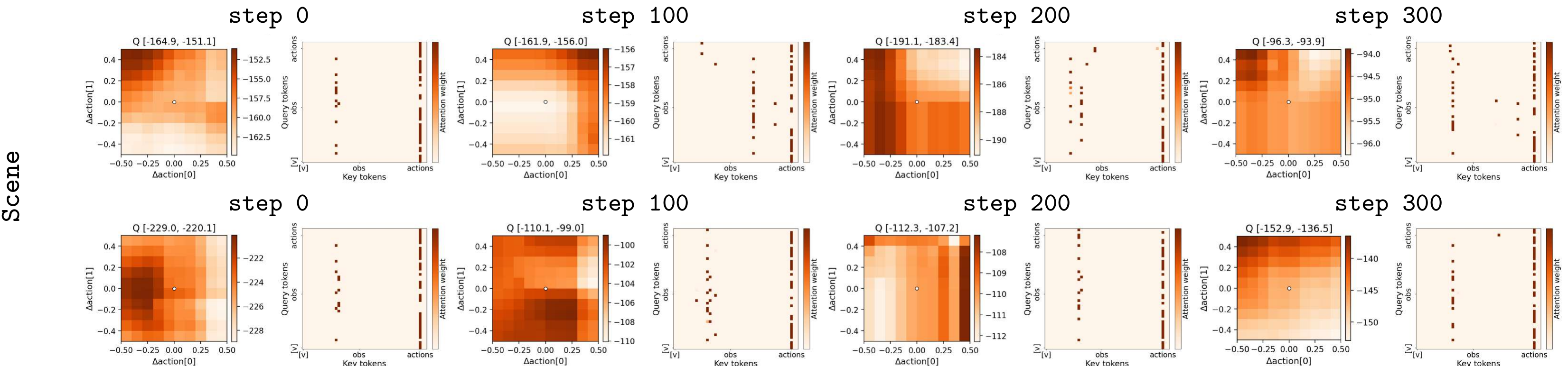}
    \caption{Attention maps and Q-value landscapes of a Transformer without \ours{} across all five environments. The learned Q-value functions exhibit non-smooth landscapes, accompanied by overfitted attention patterns, which hinder effective value learning.
    } 
    \label{fig:full_aq_wo_attn}
\end{figure*}

\begin{figure*}[h!]
    \centering
    \includegraphics[width=0.99\linewidth]{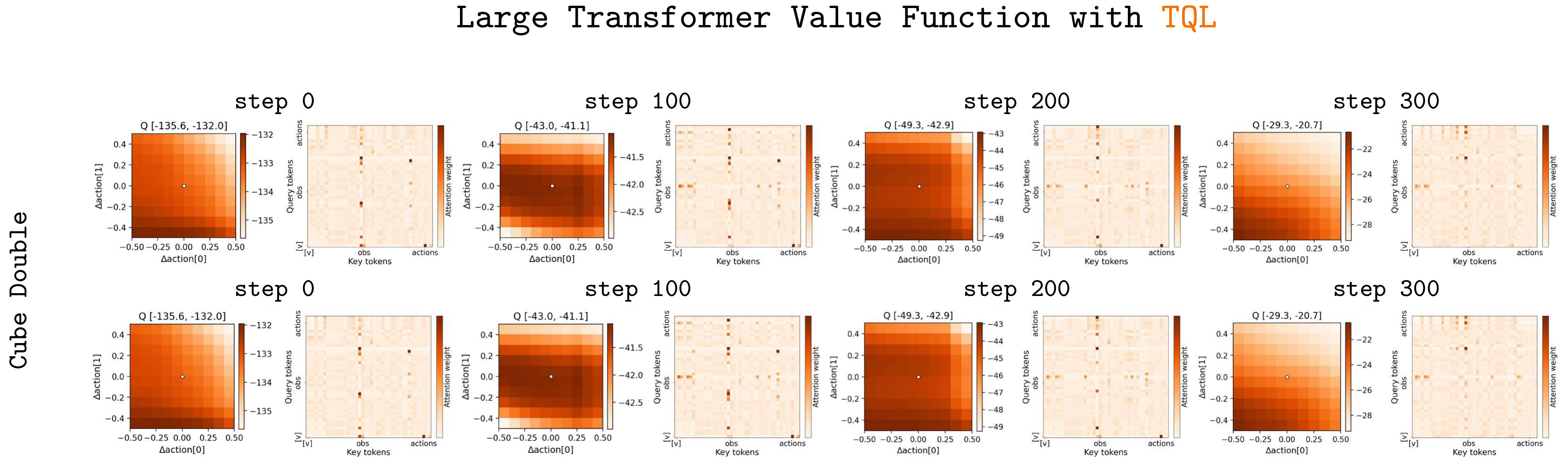}
    \vspace{1pt}
    \includegraphics[width=0.99\linewidth]{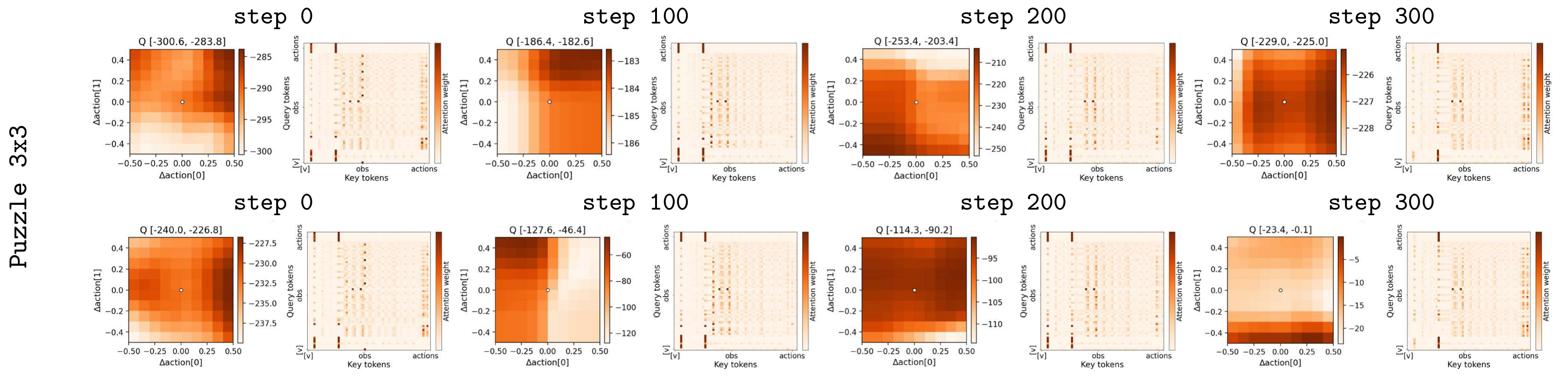}
    \vspace{1pt}
    \includegraphics[width=0.99\linewidth]{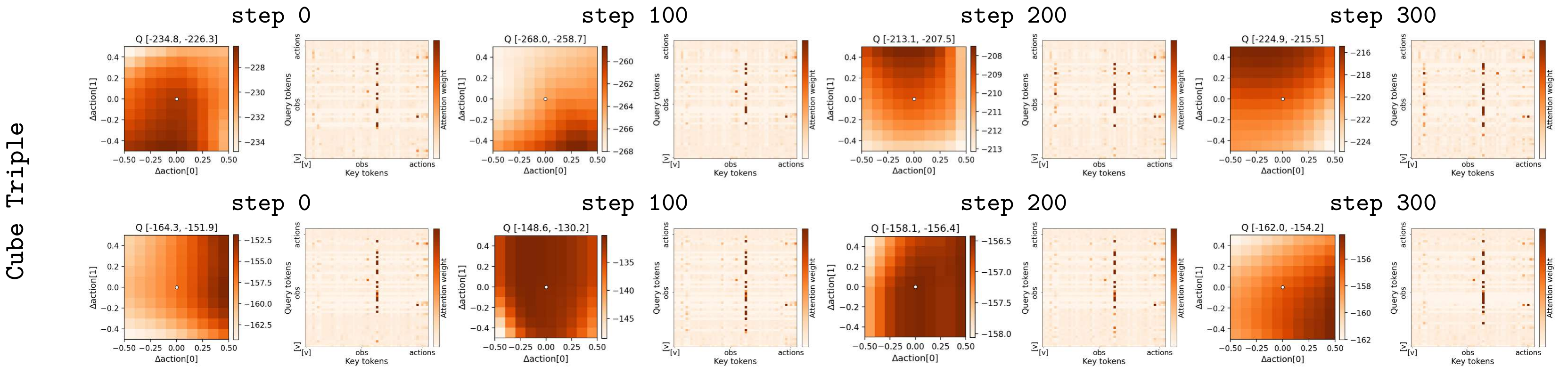}
    \vspace{1pt}
    \includegraphics[width=0.99\linewidth]{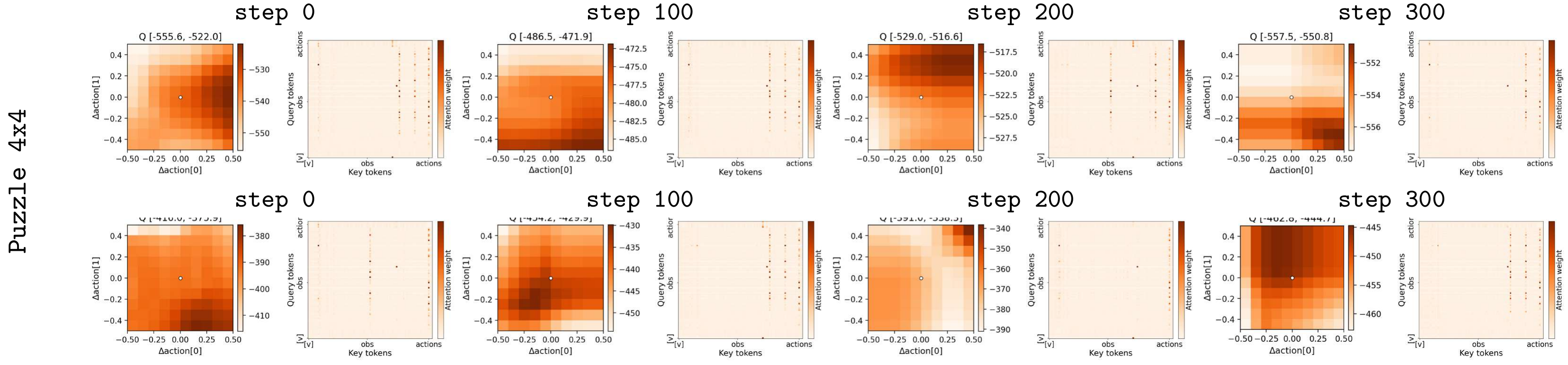}
    \vspace{1pt}
    \includegraphics[width=0.99\linewidth]{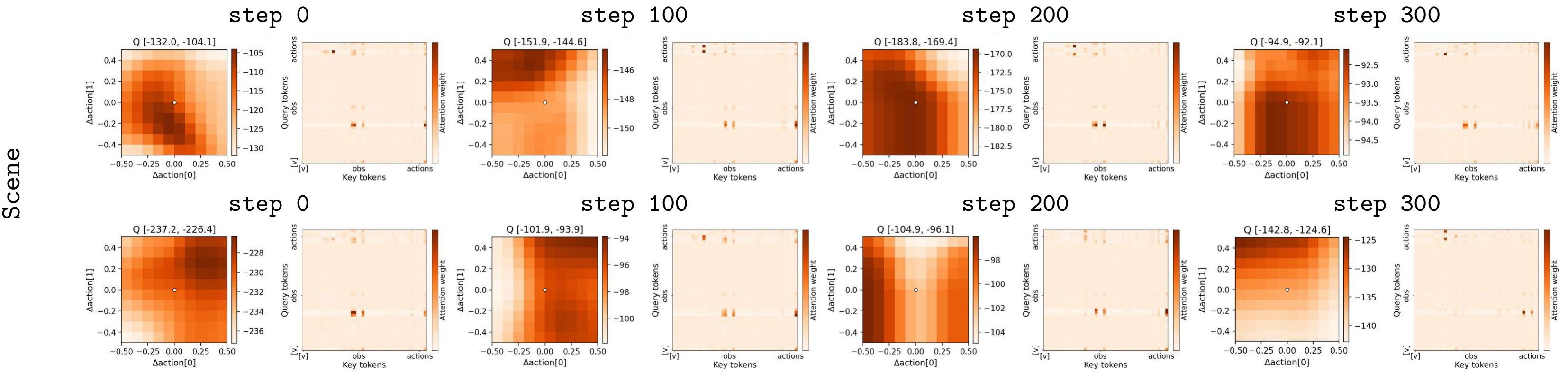}
    \caption{Attention maps and Q-value maps of \ours{} across all five environments. The learned Q-value functions exhibit smooth landscapes, accompanied by more distributed attention patterns, which facilitate stable and effective value learning.
    } 
    \label{fig:full_aq_w_attn}
\end{figure*}

\Cref{fig:full_aq_wo_attn} displays the results for the Transformer without \ours{}. We observe that the attention patterns become extremely sparse and concentrate heavily on a distinct few tokens. This collapse leads to Q-value maps that are highly discontinuous and irregular. Conversely, \Cref{fig:full_aq_w_attn} demonstrates that our training objective successfully mitigates this issue. The attention mechanism maintains a broader distribution, resulting in a significantly smoother and more consistent Q-value landscape.

\clearpage

\clearpage

\section{Additional Experiments}
\subsection{Increasing Depth}
\begin{figure}[h]
    \centering
    \includegraphics[width=\columnwidth]{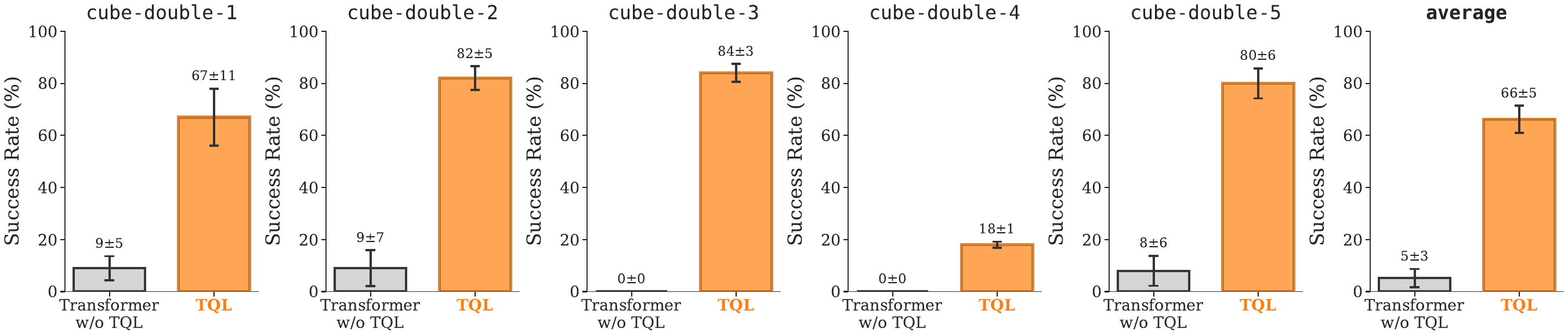}
    \caption{
        \small
        \textbf{Performance of \ours{} with more layers. } A transformer without \ours{} results in the same attention collapse as in the setting with fewer layers and \ours{} addresses this to improve performance. We find that the increased depth results in similar performance in both cases.
    }
    \label{fig:depth_result}
    \vspace{-0.5cm}
\end{figure}

\begin{figure}[h]
    \centering
    \includegraphics[width=\columnwidth]{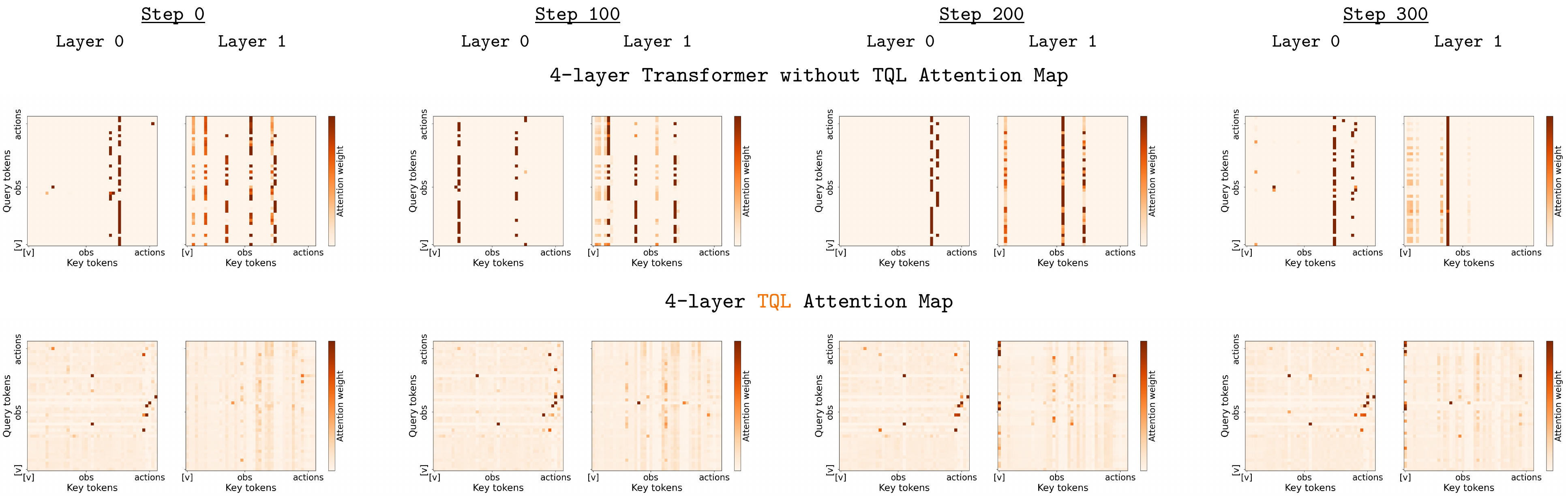}
    \caption{
        \small
        \textbf{Attention Map of \ours{} with increased depth. } We find that the attention collapse without TQL and the more uniform attention scores with TQL are consistent, regardless of depth.
    }
    \label{fig:depth_attn_vis}
\end{figure}

In our primary experiments, we scale the transformer network by increasing the hidden dimension. In this section, we show another common kind of scaling for the model described in~\cref{sec:setup} as well as \ours{}: increasing the depth of the network. To do so, we run a parameter-controlled experiment, fixing the total number of parameters at our largest setting of around 26M. We scale to a depth of 4 layers and pick the hidden dimension such that the total parameters match the 26M-parameter setting as closely as possible. As in our primary scaling experiments in \cref{ssec:scaling_exp}, we follow standard practice in keeping a fixed hidden dimension per attention head, in our case 32. We present the results in Figure~\ref{fig:depth_result}, attention maps in Figure~\ref{fig:depth_attn_vis}. As we see from the results, increasing depth for the baseline transformer results in the same entropy collapse and poor performance as in the case of increasing hidden dimensions, which is expected as increasing depth also increases capacity similar to increasing width. As in the width setting, \ours{} addresses this by preventing entropy collapse and resulting in increased performance. In fact, \ours{} achieves very similar performance with 2 layers and 4 layers (67\% vs. 66\%), suggesting that total network size is a good indicator of performance.

\subsection{Comparing to Approaches for Stabilizing Transformers in Supervised Learning} 

In supervised learning, numerous approaches have been proposed to improve the stability of training large-scale transformers. These approaches are typically intended to allow practitioners to push the stability boundary of training, for example to train at higher learning rates~\cite{rybakov2024methodsimprovingllmtraining}, whereas in stable regimes, these methods are largely ineffective.

In this section, we analyze the performance of these approaches for preventing entropy collapse---and the subsequent performance drop---in the setting of training value functions, focusing on three primary approaches:


\begin{enumerate}
  \item QK Normalization~\citep{dehghani2023scaling}, a method which adds a LayerNorm after both the query and key projections
  \item $\sigma$Reparam~\citep{zhai2023stabilizing}, a method based on spectral normalization~\citep{yoshida2017spectralnormregularizationimproving} proposed to stabilize attention entropy
  \item A transformer implementation with a set of techniques used to stably train larger models which contains RMS Norm~\citep{zhang2019rootmeansquarelayer}, Sandwich Norm~\citep{ding2021cogviewmasteringtexttoimagegeneration}, QK Norm~\citep{dehghani2023scaling}, and SwiGLU~\citep{shazeer2020gluvariantsimprovetransformer}
\end{enumerate}

\begin{figure}[h]
    \centering
    \includegraphics[width=\columnwidth]{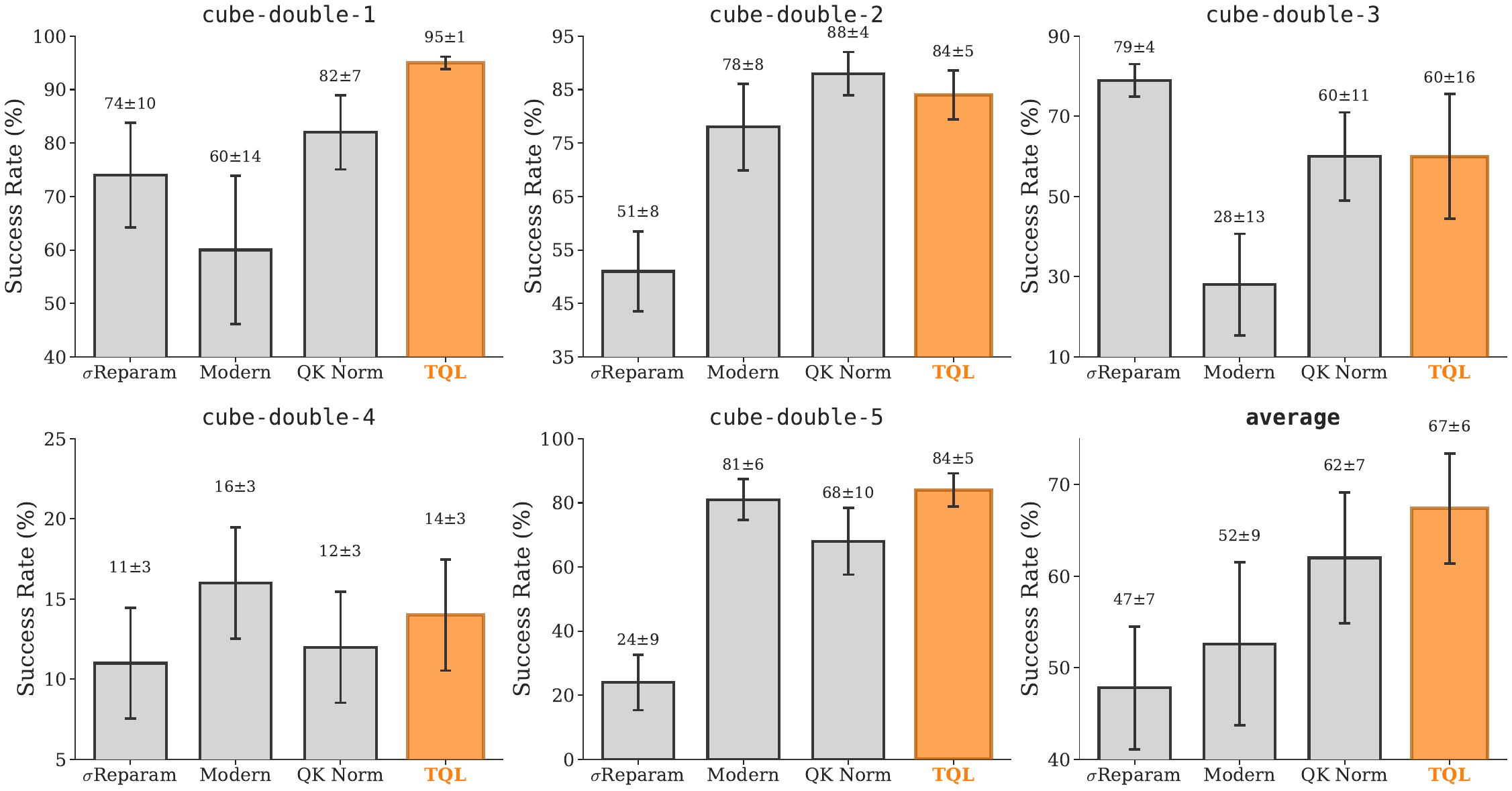}
    \caption{
        \small
        \textbf{Performance of other transformer stabilization techniques compared with \ours{}. } \ours{} achieves higher performance compared to approaches to stabilize transformer training in supervised learning.
    }
    \label{fig:norm_result}
\end{figure}

\begin{figure}[h]
    \centering
    \includegraphics[width=\columnwidth]{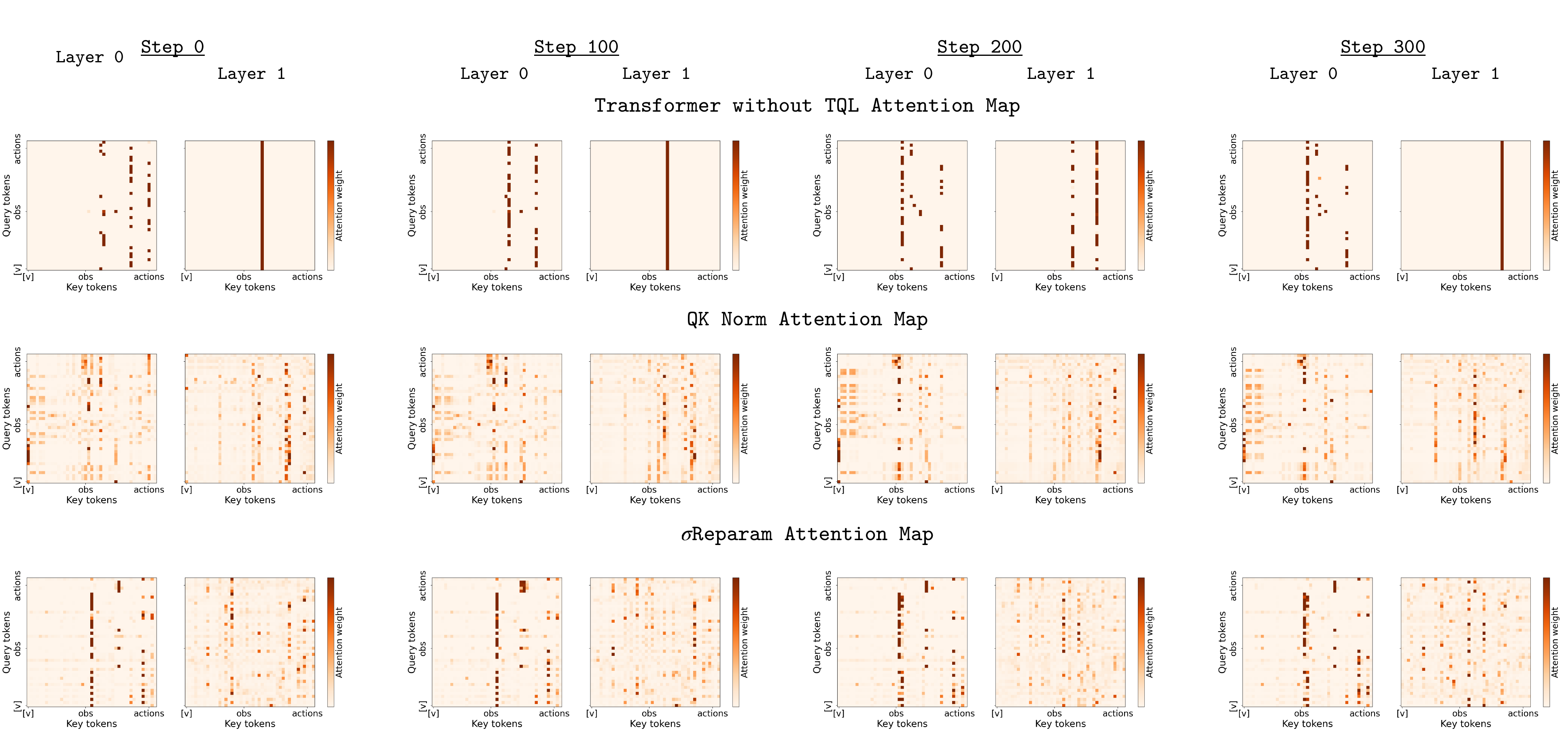}
    \includegraphics[width=\columnwidth]{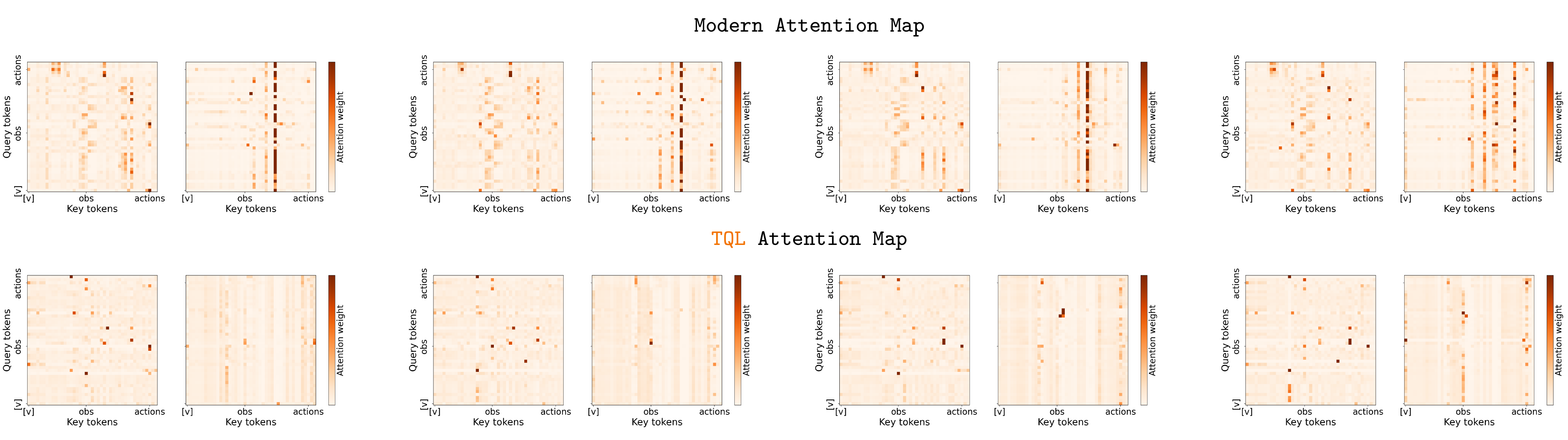}
    \caption{
        \small
        \textbf{Attention Map of other transformer stabilization techniques compared with \ours{}. } We see that common approaches for stabilizing transformers in supervised learning indeed result in higher entropy compared to the vanilla transformer. However, we see that for some layers the entropy can still collapse, whereas TQL has a more uniform attention pattern.
    }
    \label{fig:norm_attn}
\end{figure}

We present the performance in Figure~\ref{fig:norm_result} and the corresponding attention entropy in Figure~\ref{fig:norm_attn}. From the figure, we see that these approaches can help mitigate entropy collapse and result in higher entropy values and performance compared to the transformer baseline. However, these approaches are not able to consistently do so and perform worse in general compared to \ours{} when applied to value learning.



\section{Experimental Details}
\subsection{Benchmark \& Evaluation} 

We evaluate our method on the OGBench benchmark suite~\citep{park2025ogbench}. OGBench is a large-scale benchmark designed for offline goal-conditioned reinforcement learning and additionally provides single-task variants that are compatible with standard reward-maximizing RL algorithms. In this work, we adopt the single-task variants for all domains. For each task, the reward ranges from $-n_{\text{task}}$ to $0$, depending on the number of completed subtasks.

We use the following OGBench datasets for each domain:
\begin{itemize}
    \item \texttt{cube-double-play-v0}
    \item \texttt{cube-triple-play-v0}
    \item \texttt{scene-play-v0}
    \item \texttt{puzzle-3x3-play-v0}
    \item \texttt{puzzle-4x4-play-v0}
\end{itemize}

\begin{figure}[h]
    \centering
    \includegraphics[width=\columnwidth]{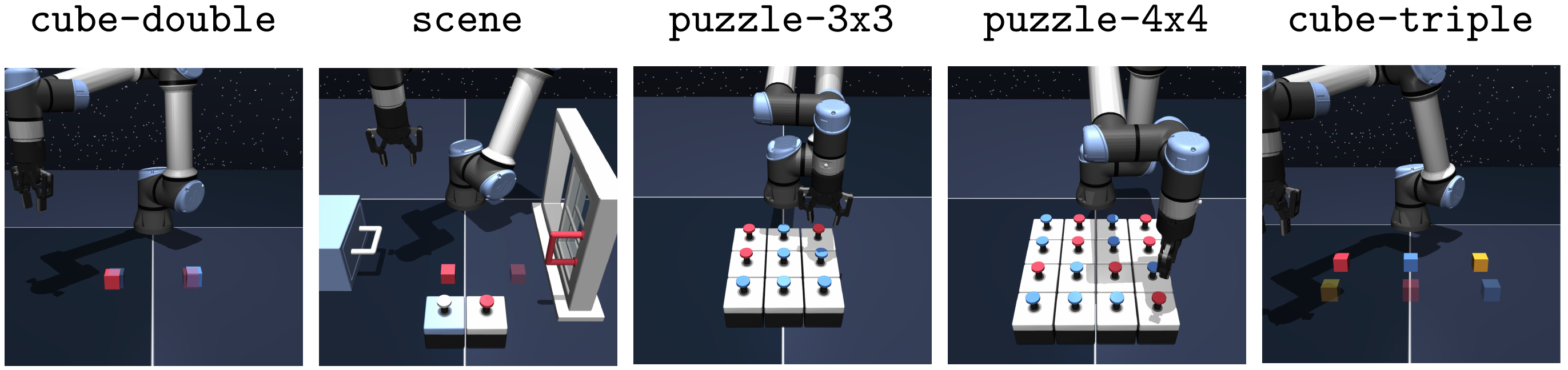}
    \caption{
        \small
        \textbf{Visualizations of the domains we evaluate on. } Each of the 5 domains from OGBench has 5 tasks, for a total of 25 tasks.
    }
    \label{fig:benchmark}
\end{figure}

The \texttt{cube-double} and \texttt{cube-triple} tasks involve complex pick-and-place manipulation of multiple colored cube blocks. The \texttt{scene} tasks require long-horizon reasoning and interaction with diverse objects in cluttered environments. The \texttt{puzzle-3x3} and \texttt{puzzle-4x4} tasks are based on the \emph{Lights Out} puzzle and are solved using a robotic arm, further evaluating the agent’s ability to generalize over combinatorial state spaces. Visualizations of all environments are shown in \cref{fig:benchmark}.

All experiments follow the official evaluation protocols and metrics defined by OGBench~\citep{park2025ogbench}. For each setting, we run three random seeds and report the mean and standard deviation of the success rate, averaged over the $800$k, $900$k, and $1$M training step checkpoints. For bar plots, we report the same average with standard error as error bars.

\label{appendix:exp_details}

\subsection{Model Structure Details}
We parameterize the Q-function \(Q(s,a)\) with a Transformer that models interactions between state and action components at the token level. Given a state \(s \in \mathbb{R}^{n_s}\) and an action \(a \in \mathbb{R}^{n_a}\), we construct a token sequence by treating each scalar dimension of the state and action as an individual token. Specifically, \(s\) and \(a\) are reshaped into sequences of length \(n_s\) and \(n_a\), respectively, and each token is independently projected into a shared hidden space of dimension \(H\) using learned linear projections. To distinguish state tokens from action tokens, we add learnable modality embeddings to the corresponding token representations. The state and action token sequences are concatenated to form a sequence of length \(n_s + n_a\).

Learnable positional embeddings are added to all tokens, and a \([{\tt value}]\) token is prepended to the sequence to aggregate global state-action information. The resulting sequence is processed by \(L\) Transformer blocks with a pre-layer normalization architecture. Each block consists of multi-head self-attention followed by a two-layer feedforward network with GELU activations, with residual connections and dropout applied after both the attention and feedforward sublayers. Following common practice, we fix the head dimension, choosing $d_h = 32$ such that the number of heads $h$ is determined by $h = d_{\text{model}}/32$. The entropy computation of \ours{} can be done either after averaging attention scores over heads, or before averaging over heads and then averaging the entropies. In the experiments in this paper, the attention is averaged over heads in the layer before computing the entropy. After a final layer normalization, the representation corresponding to the \([{\tt value}]\) token is used as a global summary of the input. Finally, we predict an ensemble of \(K\) Q-value estimates using \(K\) lightweight MLP heads applied to the \([{\tt value}]\) embedding, producing \(Q(s,a) \in \mathbb{R}^{K}\).


For the scaling experiments, we set \(L=2\) for the number of Transformer layers and vary the network size by setting the hidden dimension to \(\{128, 256, 512, 1024\}\). Unless otherwise specified, we report the final performance of the largest model with hidden dimension \(1024\).

\subsection{Optimization Hyperparameters}
\label{sec:appendix_detail}
In this section, we provide additional optimization and training details for \ours{}. All hyperparameters used in our experiments are summarized in Table~\ref{tab:params}. Following ~\citet{park2025flowqlearning}, we use environment-specific BC coefficients $\alpha$ to account for differences in dataset quality and task difficulty. For the learning rate schedules, we adopt cosine decay for both the actor and critic. The actor learning rate is fixed to 5e-4, while the critic learning rate is set to 1e-4, which we found to be the best overall across dimensions, except for the largest model setting which we tune separately.

\begin{table}[!h]
\centering
\caption{Default training hyperparameters for \ours{}.}
\label{tab:params}
\begin{tabular}{ll}
\toprule
\textbf{Hyperparameter} & \textbf{Value} \\
\midrule
Optimizer & AdamW (weight decay $0.01$) \\
Training steps & $1\,\text{M}$ \\
Batch size & $256$ \\
Discount factor ($\gamma$) & $0.99$ \\
Target smoothing coefficient ($\tau$) & $0.005$ \\
Actor learning rate & 5e-4 with cosine schedule \\
Critic learning rate & 1e-4 with cosine schedule \\
BC coefficient ($\alpha$) & Environment-specific, see Table~\ref{tab:hyperparameters} \\
Target attention entropy ($\bar{H}$) & Environment-specific, see Table~\ref{tab:hyperparameters} \\
Actor flow steps & $10$ \\
Number of action samples & $16$ \\
\bottomrule
\end{tabular}
\label{tab:hyperparams}
\end{table}

\textbf{Attention Entropy Target. } 
The attention entropy target $\bar{H}$ is an important hyperparameter in \ours{}. The scale of attention entropy is primarily determined by the input dimensionality of the task, specifically the number of state dimensions $n_{\text{s}}$ and action dimensions $n_{\text{a}}$. In our experiments, we observe that the optimal target entropy depends on the environment being evaluated and is largely independent of the model size, but the entropy target should be tuned for each problem and setting for best performance.  

To initialize $\bar{H}$ for each environment, we first compute a coarse upper bound on the attention entropy based on the input dimensionality. Concretely, the maximum entropy is 
\begin{equation}
H_{\max} = \ln\left(1 + n_{\text{s}} + n_{\text{a}}\right),
\end{equation}
and set the initial target entropy to $0.8 \times H_{\max}$. Starting from this value, we perform a local hyperparameter search within a $\pm 0.5$ range.

We further find that assigning smaller target entropy to the output layers leads to more stable training and encourages more deterministic predictions. We set a fixed $-0.5$ lower target entropy for the output layer in our experiments. All entropy tuning experiments are conducted using the second-largest model configuration, and the selected target entropy values are then fixed and reused for all model sizes within the same environment. The final target attention entropy values used for each environment are reported in Table~\ref{tab:hyperparameters}.

\subsection{Baseline Implementation Details}
\label{appendix:baseline}

We summarize the implementation details of all baseline methods used in our experiments. For BC, IQL, ReBRAC, FBRAC and IFQL we directly report the results from prior work~\citep{park2025flowqlearning,dong2025valueflows}, which are averaged over 8 seeds. We report results of floq from their provided Weights and Biases logs, which are over 3 seeds. Domain-specific hyperparameters for all methods are summarized in \cref{tab:hyperparameters}. We report results for the $\sim$26M parameters network size setting for all transformer-based methods to ensure a fair comparison. Below, we provide additional implementation details for the remaining baselines.

\paragraph{FQL~\citep{park2025flowqlearning}.}
For FQL, we report the results from the original paper for the main comparisons. For the scaling experiments, we train the critic with a fixed depth of four layers while varying the hidden dimension over \{320, 512, 1536, 2944\} to study the effect of model capacity. All hyperparameter settings are kept the same as in the official implementation.

\paragraph{floq~\citep{agrawalla2025floqtrainingcriticsflowmatching}.}
For floq, we report performance at $1$M training steps from the results provided in their released Weights and Biases logs, to ensure consistency in training steps with other baselines and our method. In the scaling experiments, we train the floq critic with four layers and vary the hidden dimension over $\{320, 512, 1536, 2944\}$. All hyperparameter settings are kept the same as in the official implementation.

\paragraph{Q-Transformer (Q-T)~\citep{chebotar2023qtransformer}.}
For Q-Transformer, we use a causal Transformer decoder architecture as described in~\citep{brohan2023rt1roboticstransformerrealworld}. We implement QK Norm~\cite{dehghani2023scaling} to reduce attention collapse, and we use learned positional embeddings as the sequence length is fixed. With the exception of the causal mask, QK-Norm, and the input and output projections, this closely matches our baseline architecture, allowing for a direct comparison. We sweep over action bins, learning rate, AdamW weight decay, and the conservatism weight, finding an optimal $N=64$, $lr=2e-5$, $\lambda=0.02$, and $\alpha=0.5$ (as defined in Eq. 2 of \citet{chebotar2023qtransformer}). We find that such comparatively low learning rates and high decay are necessary to achieve reasonable performance and otherwise observe significant instability. Moreover, without QK Norm we observe significant entropy collapse.

\paragraph{Perceiver Actor-Critic (PAC)~\citep{springenberg2024offlineactorcriticreinforcementlearning}.} For PAC, we follow the critic architecture specified in the original paper, using $32$ latent tokens and discretizing the value space into bins. A dense encoder is employed for both state and action representations. For value discretization, we set the lower bound to the minimum possible return and the upper bound to $0$ for each task. The number of value bins is chosen such that each bin has width $1$, corresponding to the minimum reward difference across all evaluated tasks. Although PAC is originally designed to jointly learn discrete action prediction, we find that this component leads to unstable training and worse performance in our evaluation setting. To ensure a fair comparison, we disable discrete action prediction and use the same policy extraction mechanism~\citep{park2025flowqlearning} as \ours{} to focus on value learning. For the scaling experiments, we train PAC models with two hidden layers and vary the hidden dimension over $\{80, 160, 320, 640\}$.

\begin{table}[H]
\caption{\footnotesize \textbf{Domain specific hyperparameters for \ours{} and baselines. For attention entropy target $\bar{H}$}, the first dimension corresponds to Transformer layers. Within each tuple, the first value specifies the target entropy for the \textit{value token}, while the second value applies to all other tokens.}
\label{tab:hyperparameters}
\begin{center}
\setlength{\tabcolsep}{4pt}
\scalebox{0.9}{
\centering
\begin{tabular}{lcccccccccc}
\toprule
& IQL & \multicolumn{2}{c}{ReBRAC} & FBRAC & IFQL & FQL & floq & PAC & \multicolumn{2}{c}{\ours{}} \\
\cmidrule(lr){2-2} \cmidrule(lr){3-4} \cmidrule(lr){5-5} \cmidrule(lr){6-6} \cmidrule(lr){7-7} \cmidrule(lr){8-8} \cmidrule(lr){9-9} \cmidrule(lr){10-11}
Domain or task & $\alpha$ & $\alpha_1$ & $\alpha_2$ & $\alpha$ & $N$ & $\alpha$ & $\alpha$ & $\alpha$ & $\alpha$ & $\bar{H}$ \\
\midrule
\texttt{cube-double-play} &
$0.3$ & $0.1$ & $0$ & $100$ & $32$ & $300$ & $300$ & $300$ & $300$ &
$((3.0,\,3.0),\,(2.5,\,2.5))$ \\

\texttt{cube-triple-play} &
$10$ & $0.03$ & $0$ & $100$ & $32$ & $300$ & $300$ & $300$ & $300$ &
$((3.5,\,3.5),\,(3.0,\,3.0))$ \\

\texttt{puzzle-3x3-play} &
$10$ & $0.3$ & $0.001$ & $100$ & $32$ & $1000$ & $1000$ & $1000$ & $1000$ &
$((3.5,\,3.5),\,(3.0,\,3.0))$ \\

\texttt{puzzle-4x4-play} &
$3$ & $0.3$ & $0.01$ & $300$ & $32$ & $1000$ & $1000$ & $1000$ & $1000$ &
$((3.5,\,3.5),\,(3.0,\,3.0))$ \\

\texttt{scene-play} &
$10$ & $0.1$ & $0.001$ & $100$ & $32$ & $300$ & $300$ & $300$ & $300$ &
$((3.5,\,3.5),\,(3.0,\,3.0))$ \\
\bottomrule
\end{tabular}
}
\end{center}
\end{table}

\end{document}